\documentclass[journal]{IEEEtran}
\usepackage{graphicx}
\usepackage{cite}
\usepackage{picinpar}
\usepackage{amsmath}
\usepackage{url}
\usepackage{flushend}
\usepackage{algpseudocode}
\usepackage{colortbl}
\usepackage{soul}
\usepackage{multirow}
\usepackage{pifont}
\usepackage{color}
\usepackage{alltt}
\usepackage[hidelinks]{hyperref}
\usepackage{enumerate}
\usepackage{siunitx}
\usepackage{amsfonts}
\usepackage{breakurl}
\usepackage{epstopdf}
\usepackage{pbox}
\usepackage{graphicx} 
\usepackage{float} 
\usepackage{CJK}
\usepackage[top=2cm, bottom=2cm, left=2cm, right=2cm]{geometry}
\usepackage{algorithm}
\usepackage{algorithmicx}
\usepackage{algpseudocode}
\usepackage{amsmath}
\usepackage{xcolor}
\usepackage{colortbl,booktabs}
\usepackage{amsmath}
\usepackage{caption}
\usepackage{tabularx}
\usepackage[tableposition=top]{caption}
\usepackage{subcaption}
\usepackage{lscape}
\begin{document}
\title{Cost-sensitive Boosting Pruning Trees for depression detection on Twitter}

\author{
	\vskip 1em
	{Lei Tong, Zhihua Liu, Zheheng Jiang , Feixiang Zhou, Long Chen, Jialin Lyu, Xiangrong Zhang \IEEEmembership{Senior Member,~IEEE}, Qianni Zhang, Abdul Sadka \IEEEmembership{Senior Member,~IEEE}, Yinhai Wang, Ling Li and Huiyu Zhou
	\thanks{L. Tong, Z. Liu, Z. Jiang, F. Zhou, L. Chen, J. Lyu and H. Zhou are with School of Computing and Mathematical Sciences, University of Leicester, United Kingdom. E-mail: \{lt228;zl208;zj53;fz64;lc408;jl766;hz143\}@leicester.ac.uk. H. Zhou is corresponding author.}
	\thanks{X. Zhang is with the Key Laboratory of Intelligent Perception and Image Understanding of Ministry of Education of China, Xidian University, China. E-mail: xrzhang@mail.xidian.edu.cn.}
	\thanks{Q. Zhang is with School of Electronic Engineering and Computer Science, Queen Mary, University of London, United Kingdom. E-mail: qianni.zhang@qmul.ac.uk.}
	\thanks{A. Sadka is with Centre for Media Comms Research, Brunel University London, United Kingdom. E-mail: abdul.sadka@brunel.ac.uk.}
\thanks{Y. Wang is with Discovery Sciences, AstraZeneca R$\&$D, Darwin Building, Cambridge Science Park, Milton Road, Cambridge CB4 0WG. E-mail: yinhai.wang@astrazeneca.com.}
	\thanks{L. Li is with School of Computing, University of Kent. E-mail: C.Li@kent.ac.uk.}
\thanks{Manuscript received on August 2020; revised xxxx.}}
	
	}

\maketitle
\begin{abstract}
Depression is one of the most common mental health disorders, and a large number of depressed people commit suicide each year.  Potential depression sufferers usually do not consult psychological doctors because they feel ashamed or are unaware of any depression, which may result in severe delay of diagnosis and treatment. In the meantime, evidence shows that social media data provides valuable clues about physical and mental health conditions. In this paper, we argue that it is feasible to identify depression at an early stage by mining online social behaviours. Our approach, which is innovative to the practice of depression detection, does not rely on the extraction of numerous or complicated features to achieve accurate depression detection. Instead, we propose a novel classifier, namely, Cost-sensitive Boosting Pruning Trees (CBPT), which demonstrates a strong classification ability on two publicly accessible Twitter depression detection datasets. To comprehensively evaluate the classification capability of CBPT, we use additional three datasets from the UCI machine learning repository and CBPT obtains appealing classification results against several state of the arts boosting algorithms. Finally, we comprehensively explore the influence factors for the model prediction, and the results manifest that our proposed framework is promising for identifying Twitter users with depression.

\end{abstract}


\markboth{IEEE Transactions on Affective Computing}%
{}

\definecolor{limegreen}{rgb}{0.2, 0.8, 0.2}
\definecolor{forestgreen}{rgb}{0.13, 0.55, 0.13}
\definecolor{greenhtml}{rgb}{0.0, 0.5, 0.0}

\section{Introduction}
Depression is one of the most common mental illnesses. It is estimated that nearly 360 million people suffer from depression \cite{cheng2006genome}.In Britain, 7.8\% of people meet the criteria of depression diagnosis, and 4-8\% will experience depression in their lifetime \cite{mcmanus2009adult}.
Andrade et al. \cite{andrade2003epidemiology} reported that the probability for an individual to encounter a major episode of depression within a period of one year is 3-5\% for males and 8-10\% for females. Because of depression, about one million of people committed suicide annually in the world \cite{cheng2006genome}.

Depressed people may have a variety of symptoms: having troubles in going to sleep or sleeping too much, lacking of passion or feeling disappointed \cite{radloff1977ces}. In clinical exercises,  psychological specialists are looking for reliable methods to detect and prevent depression. Yang et al.\cite{yang2013detecting} investigated the relation between vocal prosody and changes in depression severity over time. Alghowinem et al. \cite{alghowinem2016multimodal} examined human behaviours such as speaking behaviours and eye activities associated with major depression. Diagnostic and Statistical Manual of Mental Disorders \cite{shen2017depression} is an important reference for psychological doctors to diagnose depression. There are nine classes of depression symptoms recorded in the menu, describing the distinguishing behaviours in our daily life. Nevertheless, the symptoms of depression disorders evolve over time and it has been advised to dynamically update the criteria of depression diagnosis \cite{cheng2006genome}.

On the other hand, depression sufferers who do not receive timely psychotherapy will develop worse conditions. More than 70\% of people in the early stage of depression do not consult psychological doctors, and their conditions deteriorated \cite{shen2017depression}.  Gonz{\'a}lez-Ib{\'a}nez et al. \cite{gonzalez2011identifying} reported that people are somehow ashamed or unaware of depression which makes them miss timely treatment. Choudhury et al. \cite{de2013predicting} and Neuman et al.\cite{neuman2012proactive} proposed to explore the correlation of depression sufferers with their online behaviours on social networks.
\begin{figure*}
\includegraphics[width=1\textwidth]{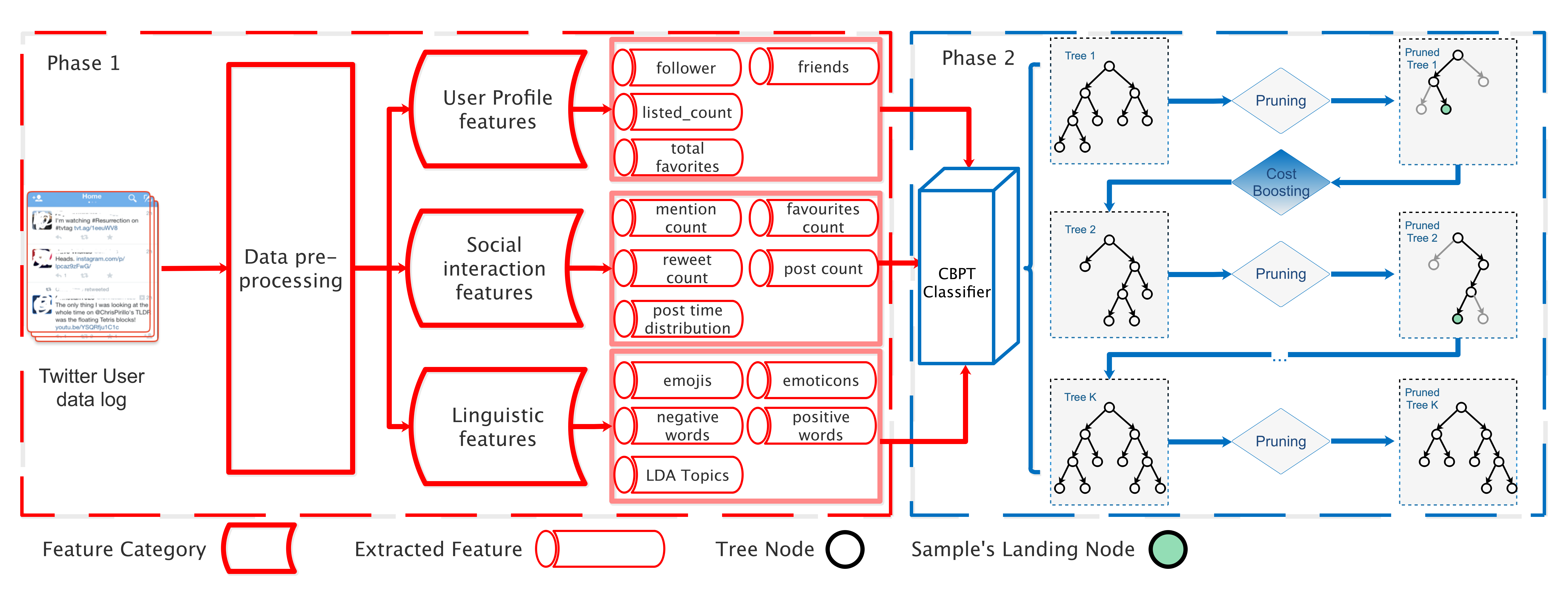}
\caption{Proposed framework. In phase 1, we conduct data preprocessing and extract various discriminative features of Twitter users. In phase 2, the CBPT classifier combines the power of $K$ pruned trees. The cost-sensitive boosting structure relies on the landing position of samples in the pruned tree structure and an example of the sample decision path is highlighted in dark black in the diagram. }
\label{fig:system_framework}
\end{figure*}
With the explosive growth of computer network applications, social networks have become an indispensable part of many people's daily lives. 62\% of the American adults (age 18 and older) use Facebook, whilst the majority of the users (70\%) visit Internet daily and a large portion of the users access to Internet multiple times each day \cite{zaydman2017tweeting}. There are 1.10 billion posts on Facebook every day. Twitter and Tumblr also have 500 and 77.5 million users who are active per day, where 70\% of the Twitter users log in every day \cite{zaydman2017tweeting}. Therefore, social networks provide a means for capturing behavioural attributes that are relevant to an individual's thinking, mood, communication, activities and socialisation\cite{de2013predicting}. Research studies reveal that collecting social networking information for analysing human physical and mental wellness is possible \cite{akbari2016tweets,coppersmith2014quantifying}. Neuman et al. \cite{neuman2012proactive} developed working methods for recognising associated signals in the user's posts on social networks, which suggest whether or not clinical diagnosis is required, based on his/her naturally occurring linguistic behaviours. Salawu et al. \cite{salawu2017approaches} detected cyber-bullying on social networks by comparing textual data against the identified traits. Nguyen et al.\cite{nguyen2014affective} utilised psycholinguistic clues to conduct sentiment analysis on users' posts to detect depressed users online. Hence, it is feasible to detect depression via social networks.

Our proposed framework is shown in Fig. \ref{fig:system_framework}. In the first phase, we conduct data preprocessing and extract discriminative features from the posts of Twitter users, while the second phase presents a new Cost-sensitive Boosting Pruning Trees method based on the Discrete Adaboost \cite{freund1996experiments} to classify the users. Our new contributions reported in this paper are:
\begin{enumerate}[(1)]

\item {We propose a novel resampling weighted pruning algorithm which dynamically determines optimal depths/layers and leaves of a tree model. The pruning procedure can support the boosting training and improve the robustness of the base tree estimator.}
\item {We combine the proposed pruning process with a novel cost-sensitive boosting structure within an ensemble framework, namely Cost-sensitive Boosting Pruning Trees (CBPT). By introducing cost items into the learning procedure of the boosting paradigm, we highlight the uneven identification importance among the samples so that the boosting paradigm intentionally biases the learning towards the samples associated with higher identification importance.}

\item {We conduct comprehensive experiments to justify the significance of our proposed framework against two Twitter depression detection datasets, i.e. Tsinghua Twitter Depression
Dataset (TTDD) and CLPsych 2015 Twitter Dataset (CLPsych2015). The experimental results demonstrate that the prediction results are explainable against the ground-truth and our proposed framework can effectively identify Twitter users with depression.}
\end{enumerate}

\section{Related Work}
In the literature, questionnaire or online interview is one of the common means used in depression diagnosis. Lee et al. \cite{lee2000development} investigated whether or not interviewees have depressive trends using a choice questionnaire. Park et al.\cite{park2013perception} conducted a face-to-face interview with 13 active Twitter users to explore their depressive behaviours. These questionnaires and interviews have several limitations. For example, they are time-consuming and hard to be generalised. On the other hand, because of the explosive growth in the popularity of social networks, online depression detection has attracted large interests in recent years.

Many research studies for online depression detection have focused on feature detection. Choudhury et al. \cite{de2013predicting} introduced measures (e.g. egocentric social graphs and description of anti-depressant medications) to quantify the online behaviors of an individual for a year before s/he reports the onset of depression. Park et al. \cite{park2012depressive} explored the use of languages in describing depressive moods using real-time moods captured from Twitter users. Saha et al. \cite{hussain2003framework} analysed the content information of depressed users' posts by extracting topical features. Most recently, Shen et al. \cite{shen2017depression} extracted six groups' features such as user profile and engagement with online application programming interface (API) to interpret the online behaviours of depression users. However, most previous research studies focus on exploring new features of depression behaviours whilst ignoring the fitness of the classification models.

Shen et al.\cite{shen2017depression} presented a multi-modal depressive dictionary learning model (MDDL) which combines sparse dictionary learning with logistic regression to identify depressed users. Nadeem et al.\cite{nadeem2016identifying} conducted experiments to classify Major Depression Disorder (MDD) using four binary classifiers, e.g. decision tree and naive bayes. Also, Choudhury et al. \cite{de2013predicting} and Shuai et al. \cite{shuai2018comprehensive} proposed a depression detection framework based on support vector machine. Nevertheless, these established classifiers cannot achieve consistent performance due to noise or errors in the data. 

In recent years, deep learning based methods for online depression detection attracted the attention of researchers. For example, Shen et al. \cite{shen2018cross} proposed a cross-domain depression detection framework which transfers the knowledge of Twitter to classify the instances of Weibo. Their proposed method aims to improve the recognition performance in the poorly labeled target domain (Weibo) utilising the rich data of the source domain (Twitter). Ray et al. \cite{ray2019multi} proposed a multi-level attention network that combines the text, audio and video features to classify depressed people. Gamaarachchige et al. \cite{gamaarachchige2019multi} proposed a multi-task, multi-channel and multi-input framework that fuses multiple input features (e.g. emotion labels, tokens) and learns knowledges from multiple classification tasks. Their proposed method achieved good performance in the CLPsych 2015 dataset \cite{coppersmith-etal-2015-clpsych}. Orabi et al.  \cite{orabi2018deep} proposed a word embedding optimisation method which combines multiple word embedding features (e.g. Skip-Gram, CBOW). They used this technique to extract text features from the Twitter users' posts and identified the depressed users. These deep learning based method can achieve promising performance on depression detection especially on multi-level feature fusion and knowledge transfer. However, these methods lack clear interpretations to the model predictions as of which specific factor influences the predicted depression risk.

Decision trees based ensemble learning brings up the possibility of developing a powerful and interpretable model. Decision trees can reveal the feature effects to the prediction and ensemble learning uses multiple learning algorithms to obtain better predictive performance than that of using any of the constituent learning algorithms alone \cite{molnar2020interpretable,ignatov2017decision,leshem2005improvement,james2013introduction}. Our framework is based on Adaboost which is one of the typical ensemble meta-algorithms to reduce biases and variances in supervised learning \cite{breiman1996bias}. In general, Adaboost employs decision dump as its base estimator. However, decision dump cannot fit well the training data because of its simple structure. Adaboost with decision dump does not perform well in complex datasets \cite{leshem2005improvement}. Boonyanunta et al. \cite{boonyanunta2003improving} proposed a method to improve Adaboost's performance by averaging the estimators' weights or reordering estimators. Based on Adaboost, Friedman et al. \cite{friedman2001greedy} reported Gradient Boost Decision Trees (GBDT) which is the generalisation of boosting to arbitrary differentiable loss functions. Unfortunately, GBDT can be over-fitting if the data is noisy and the training process of GBDT is time consuming. Chen et al. \cite{chen2016xgboost} introduced an advanced Gradient Boost algorithm (called `XGboost') based on GBDT in 2016. Although XGboost is more flexible and efficient than GBDT, it has many parameters that are hard to tune.

In this paper, we propose a novel classification algorithm based on Adaboost that can mitigate the influence of noise or errors and have a strong fitness and generalisation ability. We introduce the details of the proposed algorithm in Section 4. In addition, we summarise the discussed classification methods in Table S4, Supplementary A.

\section{Data Preprocessing and feature extraction}
In this paper, we intend to analyse depression users' online behaviours. As the scripts on social networks may be random and unpredictable, features with different noise may be obtained and influence the detection accuracy. Before feature extraction is implemented, we carry out the following preprocessing procedure: (1) Minimisation of the influence of noisy samples. Inspired by the work of Yazdavar et al.\cite{yazdavar2017semi}, we remove the noisy samples from the dataset where the posting number of the samples is less than five. These samples cannot provide sufficient information for analysing the users' behaviours or topic modelling. (2) Processing of irregular words. The words on social networks may look irregular because of mistaken spelling or abbreviations. We use the Textblob API reported in \cite{loria2014textblob} (commonly used in natural language processing tasks) to remedy the wrong type of words. (3) Stemming. We expect to perform statistical analysis on commonly used words of control and depressed users separately and conduct topic modelling on the users' posts. Words must be of unified representations regardless of tense and voice. Hence, we utilise the SnowballStemmer algorithm reported in \cite{porter2001snowball} to deal with these words. For instance, ``accepting" and ``accepted" can be converted to ``accept". Afterwards, we extract three feature categories as follows and the proposed framework is shown in Phase 1 of  \figurename{\ref{fig:system_framework}}.

(1) User's Profile Features: The user's profile features contain the user's individual information on social networks. We collect 4 different features here: $total\_favourites$ reflects the number of posts that this particular user favours during his/her account's lifetime; $listed\_count$ shows the number of the public list that this user holds a membership within. We collect the number of the user's $friends$ and $followers$ which well characterise the author's egocentric social networks.

(2) Social Interaction Features: Park et al. \cite{park2012depressive} discovered that depressed users are less active in social networks, and depressed users regard social networking as a tool for social awareness and emotional interaction. Thus, we extract $retweet\,count$, $mention\,count$ (e.g. @someone) and $favourites\,count$ (indicating how many times this post has been favoured by the other users) to describe the behaviours of the user interacting with others. Besides, we collect the $posting\,number$ and $time\,distribution$ to demonstrate  the user's activeness on social networks.

(3) Linguistic Features: The content of the posts on social networks can intuitively reflect a person's mood and attitude. Depressed users may post more negative words than control users \cite{de2013social,park2012depressive,de2013predicting,shen2017depression}. Hence, we count the numbers of $negative$ and $positive\, words$ in the tweets using the NLTK toolkit \cite{bird2004nltk}. In addition, we collect the numbers of $emoj$i and $emoticons$ from the texts to form relevant features. In order to comprehensively explore the semantics, Resnik et al.\cite{resnik2015beyond} examined the difference of the concerned topics between depressed and control users by topic modeling and observed that topic modeling might be effective for depression detection. In our work, we utilise the Latent Dirichlet Allocation (LDA) approach presented in \cite{blei2003latent} to extract $topic\,distributions$ from the tweets.

Finally, the extracted feature sets are used to train our proposed classifier CBPT, which is shown in Phase 2 of \figurename{\ref{fig:system_framework}} and we provide the details of the extracted  feature dimensionality in Table S1, Supplementary A.
\section{Proposed Method}
\subsection{Discrete Adaboost}
Our classification algorithm is built upon the discrete Adaboost algorithm proposed by Freud et al. \cite{freund1996experiments}. Algorithm \ref{algori: standard_ada} presents the baseline scheme of the discrete Adaboost that combines many simple hypotheses (called weak learners) to form a strong classifier for the task \cite{leshem2005improvement}. The algorithm can be summarised as follows: (1) Training multiple base classifiers sequentially and assigning a weight value $ln(\beta_{m})$ according to its training error $\varepsilon_{m}$. (2) The samples misclassified by the preceding classifier are assigned a higher weight $w_{m+1,i}$, which will let the classifier pay more attention to these samples. (3) Finally, combining all the weak classifiers with their weights to obtain an ensemble classifier $G(X)$.  As we have discussed above, Adaboost may not perform well on a complex dataset, and hence we propose the CBPT algorithm to improve the performance of Adaboost in two aspects: (1) We improve the fitting and generalisation ability of the base classifier. (2) We propose a novel boosting structure to strengthen the sample re-weighting process.

 \begin{algorithm}
        \caption{Discrete Adaboost algorithm.}
        \label{algori: standard_ada}
         \textbf{Input:} A training set $D=\left \{ \left (X_{i} , y_{i} \right) \right \}_{i=1}^{N}$.\\
    	\textbf{Output:} A model $M_{K}\left(X\right)$ which is based on $K$ decision trees with their corresponding weight.
        \begin{algorithmic}[1]
        	
            		\Procedure{Adaboost}{$D$}
            	   \State {Initialise sample weight distribution $W = \left \{\left(w_{k}^{\left(i\right)}\right)\right\}$}.
			         \State {Set each sample's weight $w_{k}^{\left(i\right)}$ to $\frac{1}{N}$.}
                \For {$k \in (1,K)$}
                		\State {Fit an estimator $M_{k}\left(X\right)$ to the training data with}
                		\Statex{$\quad  \quad \quad  W_{k}$.}
                		\State {Let $u_{i}=1$ if the i-th case is classified}
                		\Statex{$\quad  \quad \quad$incorrectly, otherwise zero.}
                	 	\State {Compute training error $\varepsilon_{k}=\sum_{i=1}^{N}w_{k}^{\left(i\right)}u_{i}$.}
                		\State {Update sample's weight $ w_{k+1}^{\left(i\right)} = \frac{w_{k}^{\left(i\right)}\beta_{k}}{\sum_{i=1}^{N}w_{k}^{\left(i\right)}\beta_{k}}$,}
                		\Statex {$\quad  \quad \quad$where 	$\beta _{k} =  \frac{(1-\varepsilon_{k})}{\varepsilon_{k}}$.}
                		\State {$M_{k}\left(X\right) \leftarrow M_{k-1}\left(X\right) + \log_{e}\left ( \beta_{k} \right ) M_{k}\left(X\right)$}.
                \EndFor
                \State
                 \Return {$M_{K}\left(X\right)$}
            	\EndProcedure
        \end{algorithmic}

    \end{algorithm}

\subsection{Cost-sensitive Boosting Pruning Trees}
In this section, we propose an ensemble method that combines an improved Adaboost algorithm with pruned decision trees for classification. Here, we still employ a decision tree as the base estimator  because of its flexibility and interpretability. Decision dump often suffer from under-fitting whilst a full tree has a high variance. We here consider pruning trees in order to increase system generalisation. In our algorithm, we firstly apply all the training samples and allow a decision tree to fully grow, and then use the cost-complexity pruning method reported in \cite{breiman2017classification} to prune certain branches of the trees and use the modified criterion to evaluate the system performance with the pruned trees and update the weights. Afterwards, the above steps will be executed iteratively till the maximum number of the trees is reached.
To formulate our algorithm, we here declare the used notations in advance. In particular, we denote the training dataset as $D=\left \{ \left (X_{i} , y_{i} \right) \right \}_{i=1}^{N}$, and $X_{i}^{\left(v\right)} \in \mathbb{R}^{N \times V}$ is the sample feature vector where $N$ represents the set size and $V$ is the feature dimension. $y_{i}$ represents the training target. We employ $W = \left \{\left(w_{k}^{\left(i\right)}\right) \in \mathbb{R}^{N}\right\}_{k=1}^{K}$ to represent the set of the sample weight distribution. $K$ is the number of the estimators (iterations) and each sample's weight is initialised to $\frac{1}{N}$ in the first iteration during the normalisation. Furthermore, we use $\theta _{k}$ and $M_{K}(X)$ to denote the $k$-th estimator's weight and the ensemble classifier. 

\subsubsection{Resampling Weighted Pruning Algorithm}
In most of the previous boosting algorithms \cite{friedman2001greedy,chen2016xgboost,kokel2020unified}, except $num \  trees$, $max \ depth$ and $num \ leaves$ are two key hyperparameters which affect the classifier's performance significantly.
Manually tuning the hyperparameter combinations is a heavy task and it is hard to find the best parameter combinations for different datasets. Therefore, we propose a novel technique called resampling weighted pruning to automatically prune redundant leaves and produce robust tree models, where weights are used to establish a relationship between the pruning and boosting practice.

Firstly, we denote the original learning sample set $D$ which is divided randomly into $S$ subsets, $\left\{D_{s}\right\}_{s=1}^{S}$ and the training set of each subset is $D^{(s)}=D - D_{s}$. The tree $T_{max}$ comes from the original set $D$ and we build a complete tree on each subset $D^{(s)}$. We present the cost function of the decision trees as follows:

\begin{equation}
\label{eqn:old_cost}
\begin{aligned}
\mathcal{L}(T;w_{k}) &=\sum_{\left | \tilde{T} \right |}\left[1- \sum_{c=1}^{C} \left ( \frac{\sum_{i_{c}}w_{k}^{(i_{c})}}{\sum_{i}w_{k}^{(i)}} \right )^2\right] \\
\end{aligned}
\end{equation}
where $\left |\tilde{T} \right|$ is the leaves' number, $C$ denotes the class number and the sample of class $c$ is defined as $i_{c}$. The loss of the decision trees is the sum of all the leaf nodes' gini impurity \cite{loh2011classification}. A complete tree's loss $\mathcal{L}(T_{max};w_{k})$ is zero because each leaf node only includes a single class's samples. But $\mathcal{L}(T;w_{k})$ will increase in the pruning process where the pruned nodes are merged with their parents' nodes. Therefore, the present cost function is not a good measure of selecting a subtree because it always favours large trees. Thus, the penalty term, regularization parameter $\alpha$ and the tree leaves $\left|\tilde{T}\right|$ are added to the cost function. The new cost function is defined as follows:
\begin{equation}
\label{eqn:new_cost}
\mathcal{L}_{\alpha}(T;w_{k}) = \mathcal{L}(T;w_{k})+ \alpha   \left | \tilde{T} \right |
\end{equation}
The penalty term favours a simple tree when $\alpha$ is constant and $\left|\tilde{T}\right|$ decreases with pruning.

Now, the variation in the cost function is given by $\mathcal{L}_{\alpha}(T-T_{t};w_{k})-\mathcal{L}_{\alpha}(T;w_{k})$, where $T_{t}$ represents a branch with the node at $t$ and a tree pruned at node $t$ would be $T-T_{t}$. Next, the cost of the pruning on the internal nodes is calculated by equating $\mathcal{L}_{\alpha}\left(T-T_{t};w_{k}\right)$ to that of the branch at node $t$:

\begin{equation}
\label{eqn:variation}
\begin{aligned}
&  \mathcal{L}_{\alpha}(T-T_{t};w_{k})-\mathcal{L}_{\alpha}(T;w_{k})\leq 0\\
&\Rightarrow  \mathcal{L}_{\alpha}\left(t;w_{k}\right)-\mathcal{L}_{\alpha}\left(T_{t};w_{k}\right)\leq 0\\
&\Rightarrow  \mathcal{L}(t;w_{k}) + \alpha -\mathcal{L}(T_{t};w_{k})- \alpha \left | \tilde{T_{t}}\right |\leq 0\\
&\Rightarrow \frac{\mathcal{L}(t;w_{k})-\mathcal{L}(T_{t};w_{k})}{\left | \tilde{T_{t}}\right |-1}\leq \alpha
\end{aligned}
\end{equation}
We define:
\begin{equation}
\label{eqn:gt}
g\left(t\right) = \frac{\mathcal{L}(t;w_{k})-\mathcal{L}(T_{t};w_{k})}{\left | \tilde{T_{t}}\right |-1}
\end{equation}
We will prune branch $T_{t}$ with the decrease of the cost function value when $\alpha\geq g\left(t\right)$. The order of pruning is performed by setting $\alpha = \arg\min g(t)$ in order to find the suitable branch, which should be pruned, and the process will be repeated until the tree is left with the root node only. This provides a sequence of subtrees $\left \{(T^{(s)}_{j}); \right\}^{J}_{j=1}$ with the associated cost-complexity parameters $\left \{(\alpha_{j}); \forall \alpha \in \mathbb{R}\right\}^{J}_{j=1}$ where $J$ is the length of the subtree sequence.

For $\alpha$, we apply the pruned tree $T^{(s)}_{j}$ to predicting the estimations in the $s$-th test set, resulting in the following error rate:
\begin{equation}
\label{eqn:TE}
\mathcal{TE}^{(s)}_{j} = \frac{\sum_{i_{miss}}w_{k}^{(i_{miss})}}{\sum_{i}w_{k}^{(i)}}
\end{equation}
where $i_{miss}$ denotes the index of the misclassified sample's weight, $w_{k}^{(i)}$ is the sample's weight of the test set $D_{s}$ and $\mathcal{TE}^{(s)}_{\alpha} $ represents the misclassified rate of set $D_{s}$. Hence, the average misclassified rate of $S$ is:
\begin{equation}
\label{eqn:TE_average}
\mathcal{TE}_{j}= \frac{1}{s} \sum_{s=1}^{S} \mathcal{TE}^{(s)}_{\alpha}
\end{equation}
and we define
\begin{equation}
\label{eqn:best_alpha}
\alpha^* = \arg \min_{\alpha_{j}}\mathcal{TE}_{j}, \ \ \  \exists \alpha_{j}>0
\end{equation}
which is the best pruned tree obtained by pruning $T_{max}$ till $\mathcal{L}_{\alpha^*}(T_{max};w_{k})$ reaches the minimum. The pseudocode of our resampling weighted pruning algorithm is shown in Algorithm \ref{algori: RWPA}.

    \begin{algorithm}
        \caption{Resampling Weighted Pruning Algorithm.}
        \label{algori: RWPA}
        \textbf{Input:} A training set $D$ with corresponding weight $W_{k}$. \\
    	\textbf{Output:} A pruned tree estimator $M_{k}\left(X\right)$.
        \begin{algorithmic}[1]
            \Function {BestPrunedTree}{$D,W_{k}$}
            	   \State Randomly split the learning samples $D$ into $S$ folds,
            	   \Statex $\qquad \left\{D_{s}\right\}_{s=1}^{S}$.
            	   \State Grow a decision tree $T_{max}$ on the whole set $D$.
                \For {$s\in[1,S]$}
                    \State Fit a decision tree $T^{(s)}$ to subset $D^{(s)}$.
                    \State Generate subtree sequence $\left \{(T^{(s)}_{\alpha}); \forall \alpha \in \mathbb{R}\right\}$ by
                    \Statex $\quad \quad \quad $Eq. \eqref{eqn:variation}.
					 \State  Generate subtree sequence $\left \{(T^{(s)}_{j}); \right\}^{J}_{j=1} \leftarrow $
					 \Statex
					$
					\begin{cases}
					 & $1. Calculate$\  g\left(t\right) \ $using Eqs. \eqref{eqn:variation}-\eqref{eqn:gt}$ \\
						 & $2. Set $\alpha=\arg \min\,g\left(t\right)$ and prune the branch $T_{t} \\
						 & $3. Recursively repeat till the tree only has root nodes$
					\end{cases}	$

				   \State Calculate $\mathcal{TE}^{(s)}_{j}$ $\leftarrow$ Eq. \eqref{eqn:TE}.
                \EndFor
                \State Compute average error rate $\mathcal{TE}_{j}$ against each
                \Statex $\quad \quad$substree.
                \State $\alpha^* = \arg \min_{\alpha_{j}}\mathcal{TE}_{j}; (\exists \alpha>0)$.
                \State The best pruned tree estimator $M_{k}\left(X\right)$ $\leftarrow$
                \Statex  $\quad \quad$Prune $T_{max}$ till $\mathcal{L}_{\alpha^*}(T_{max};w_{k})$ becomes minimal.
               \State \Return{$M_{k}\left(X\right)$}.
            \EndFunction

        \end{algorithmic}
        \end{algorithm}

\subsubsection{Tree-based Cost-sensitive Boosting Structure}
As shown at steps 7 and 8 of Algorithm 1, Adaboost employs the training error $\varepsilon_{m}$ as the evaluation criterion of the base estimator's performance,  to set up the estimator's weights and update the samples' weights. All the misclassified samples receive the same weights in each iteration. In general, we assume that misclassified samples should be given different weights according to the ``hardness" of the samples - harder samples are of more weights. We now propose a novel boosting architecture namely Tree-based Cost-sensitive Boosting which utilizes the tree model to assess the “hardness” of the training samples and optimize the boosting process.

In the first step, we apply a complete decision tree to the training data $D$ and prune it in order to obtain the best tree estimator $M_{k}(X)$.  A complex tree model has more depths. Similarly, the deeper the landing node of a sample is, the harder the sample can be classified. Here, we present a new and effective depth penalty term as follows:
\begin{equation}
\label{eqn:depth_term}
\mathcal{DP}_{k}^{(i)} = \frac{\psi_{d}(\sigma^{(i)}_{k} - \min(\sigma_{k}))}{\max(\sigma_{k})-\min(\sigma_{k})} + \eta_{d}; \ \psi_{d} \in \mathbb{N}^{+} ,\eta_{d} \geq 1
\end{equation}
where $\sigma^{(i)}_{k}$ represents the landing node depth of sample $i$, $\max(\sigma_{k})$ and $\min(\sigma_{k})$ are the maximum and  minimum values in the node depth array $\sigma_{k}$. $\psi_{d}$ and $\eta_{d}$ are two hyper-parameters where $\psi_{d}$ is the percentage of data scaling, and $\eta_{d}$ is the lower limit of the penalty term. The depth penalty term is a coefficient that is multiplied with the original sample's weight to enable hard samples to gain more weights in the next iteration.

The landing node's depth can be regarded as the global evaluation of the "hardness" of a sample associated with the tree structure. In the pruning procedure, the pruned samples are included in the parent node of the pruned branch. Here, we use node impurity to represent the local evaluation of the "hardness" of a sample. For instance, when two samples land in different leaf nodes but with the same depth, the sample of low node impurity will be given more weights as the sample is separated from the most samples of the same class in the feature space. Hence, the impurity penalty term $\mathcal{IP}_{k}^{(i)}$ is defined as follows:


\begin{equation}
\label{eqn:impurity_gini}
\mathcal{\oe}_{k}^{(i)} = \frac{N-\mu_{k}^{(i)}}{2N} - E_{p(x)}^{(i)}\left [ \log q(x) \right ]\frac{\mu_{k}^{(i)}-N}{N}
\end{equation}
\begin{equation}
\label{eqn:impurity_term}
\begin{aligned}
\mathcal{IP}_{k}^{(i)} =& (\left \| \mathcal{DP}_{k}^{(i)} \right \|_{+\infty}-\left \| \mathcal{DP}_{k}^{(i)} \right \|_{-\infty}) \frac{(\oe^{(i)}_{k} - \min(\oe_{k}))}{\max(\oe_{k})-\min(\oe_{k})} \\
&+ \left \| \mathcal{DP}_{k}^{(i)} \right \|_{-\infty}
\end{aligned}
\end{equation}
Eq. \eqref{eqn:impurity_gini} is an inverse transformation of the impurity value, where $\mu_{k}^{i}$ is the sample number in the landing node, $E_{p(x)}^{(i)}\left [ \log q(x) \right ]$ is the impurity value either Cross Entropy or Gini Impurity, $p(x)$ and $q(x)$ are the predicted probability distributions of the sample $X_{i}$. Similarly, in Eq. \eqref{eqn:impurity_term}, we employ the data scaling for $\mathcal{\oe}_{k}^{(i)}$ and obtain the impurity penalty term $\mathcal{IP}_{k}^{(i)}$, $\left \| \mathcal{DP}_{k}^{(i)} \right \|_{+\infty}$ and $\left \| \mathcal{DP}_{k}^{(i)} \right \|_{-\infty}$ are positive and negative infinity norms of the depth penalty vector which are used to limit the range of data scaling.

The proposed two penalty terms mainly rely on the landing positions of the samples in the  pruned tree structure. Algorithm \ref{algori: treerecurse} returns the learning sample position by recursively following the sample's decision path as follows:
\begin{algorithm}
        \caption{Recursively Find Landing Node.}
        \label{algori: treerecurse}
              \textbf{Input:} Node id $l$, a learning sample $(X_{i},y_{i})$ \\
    	\textbf{Output:}  Node depth $\sigma_{k}^{(i)}$, node sample number $\mu_{k}^{(i)}$
        \begin{algorithmic}[1]

            \Function{TreeRecurse}{$l,(X_{i},y_{i})$}   \Comment{Find the tree node where the sample land }
            \If {$Node_{l}==Leaf$} \Comment{ Check if node $l$ is a leaf }
           		\State \Return $\sigma_{k}^{(i)}$, $\mu_{k}^{(i)}$
           	\Else
           		\If {$X_{i}^{(v)} < Threshold^{(v)}$ } \Comment{ Determine if the sample flow down to left or right child}
           			\State \Return TreeRecurse $\left ( a_{l} ,(X_{i},y_{i})\right )$
           		\Else
           			\State \Return TreeRecurse $\left ( b_{l} ,(X_{i},y_{i})\right )$
           		\EndIf
           \EndIf
            \EndFunction
        \end{algorithmic}

    \end{algorithm}

In each iteration $k$, the ensemble boosting aims to minimise an exponential loss function, described by:
\begin{equation}
\label{eqn:ensemble_loss}
\widetilde{L}(M)=\sum_{i=1}^{N}\ \exp[-y_{i}(M_{k-1}(X_{i})+\theta _{k}M_{k}(X_{i}))]
\end{equation}
where $M_{k-1}(X_{i})$ represent the $k-1$ trained pruned trees and $w_{k}^{(i)}= \exp(-y_{i}M_{k-1}(X_{i}))$, $\theta _{k}$ is the estimator weight of the $k$-th pruned tree. We can calculate the first order partial derivative of $\widetilde{L}(M)$ with respect to the estimator weight $\theta _{k}$:
\begin{equation}
\label{eqn:derive_loss}
\begin{aligned}
\frac{\partial \widetilde{L}(M)}{\partial \theta _{k}} &= \frac{\partial }{\partial \theta _{k}} \sum_{i=1}^{N}\ w_{k}^{(i)}\exp(-y_{i}\theta _{k}M_{k}(X_{i})) \\
&=\frac{\partial }{\partial \theta _{k}} \sum_{i|y_{i}\neq M_{k}(X_{i})}w_{k}^{(i)} e^{\theta_{k}} + \sum_{i|y_{i}= M_{k}(X_{i})}w_{k}^{(i)} e^{-\theta_{k}} \\
&=\frac{\partial }{\partial \theta _{k}} ((1-\varepsilon_{k})e^{-\theta_{k}}+\varepsilon_{k}e^{\theta_{k}}) \\
&= (\varepsilon_{k}-1)e^{-\theta_{k}}+\varepsilon_{k}e^{\theta_{k}}
\end{aligned}
\end{equation}
By taking zero to the left hand side of Eq.\eqref{eqn:derive_loss}, we have:
\begin{equation}
\label{eqn:estimator_weight}
\theta _{k} \propto \frac{1}{2}(\log \frac{(1-\varepsilon_{k})}{\varepsilon_{k}} + \log(C-1))
\end{equation}
where $\varepsilon_{k}$ is the training error of pruned tree $M_{k}(X)$. $ \log(C-1)$ is a regularisation term and $C$ is the number of classes.

The updating process of the new sample's weights is defined as:
\begin{equation}
\label{eqn: sample_weight}
\begin{aligned}
w_{k+1}^{(i)} &= \frac{\exp[-y_{i}(M_{k-1}(X_{i})+\theta _{k}M_{k}(X_{i}))]}{Z_{k+1}} \\
&= \frac{w_{k}^{(i)}}{Z_{k+1}}\exp[2\theta_{k} \log(\mathcal{DP}_{k}^{(i)})\log(\mathcal{IP}_{k}^{(i)})1_{y_{i}\neq M_{k}(X_{i})}] \\
\end{aligned}
\end{equation}
where $Z_{k}$ is a normalisation factor, and
\begin{equation}
\label{eqn: z_normalize}
\begin{aligned}
Z_{k+1} =& \sum_{i|y_{i}\neq M_{k}(X_{i})}w_{k}^{(i)}\mathcal{DP}_{k}^{(i)}\mathcal{IP}_{k}^{(i)} \frac{(1-\varepsilon_{k})(C-1)}{\varepsilon_{k}}+  \\
 &\sum_{i|y_{i}= M_{k}(X_{i})}w_{k}^{(i)}
\end{aligned}
\end{equation}
The two penalty terms are taken as the interference factors to influence the updating of the sample's weights and the misclassified samples employ different weights according to their landing positions in the pruned tree structure. The iterative training of the cost-sensitive boosting will stop if it converges (i.e. $\varepsilon_{k}$ reaches zero) or we reach the maximum iteration number $K$. The whole algorithm is illustrated in Algorithm \ref{algori: cost_boosting}, linked with the two proposed functions.
\begin{algorithm}
        \caption{ Cost-sensitive Boosting Pruning Trees Algorithm.}
        \label{algori: cost_boosting}
        \textbf{Input:} A training set $D=\left \{ \left (X_{i} , y_{i} \right) \right \}_{i=1}^{N}$ with sample distribution $W = \left \{\left(w_{k}^{\left(i\right)}\right) \in \mathbb{R}^{N}\right\}_{k=1}^{K}$\\
    	\textbf{Output:} A Cost-sensitive Boosting Pruning Trees model $M_{K}\left(X\right)$
        \begin{algorithmic}[1]
            \Procedure {CostBoosting}{$D,W$}
            	   \State {Initialize sample weight distribution $W = \left \{\left(w_{k}^{\left(i\right)}\right)\right\}$}.
			         \State {Set each sample's weight $w_{k}^{\left(i\right)}$ to $\frac{1}{N}$.}
			
                \For {$k \in (1,K)$}
                		
                		\State $M_{k}(X) \leftarrow$ BestPrunedTree$\left(D,W_{k}\right)$
                		\For {$i \in (1,N)$}
                			\State $\sigma_{k}^{(i)}$, $\mu_{k}^{(i)} \leftarrow$TreeRecurse$\left( 0, (X_{i},y_{i})\right) \ \triangleright $
                			\Statex $\quad  \quad \quad \qquad$Start from the root node.

            				\State Calculate depth penalty coefficient $DP_{k}^{(i)}$          							\Statex $\quad  \quad \quad \qquad$using Eq. \eqref{eqn:depth_term}.
            				
                			\State Calculate impurity penalty coefficient $IP_{k}^{(i)}$ 								\Statex $\quad  \quad \quad \qquad$using Eqs. \eqref{eqn:impurity_gini}-\eqref{eqn:impurity_term}.

                		\EndFor
                		\State Update the estimator weight using Eq. \eqref{eqn:estimator_weight}.

                		\State Update each sample's weight $w_{k+1}^{(i)}$ using Eq.
                		\Statex $\quad  \quad \quad$ \eqref{eqn: sample_weight}.
                           		
                		\State {$M_{k}\left(X\right) \leftarrow M_{k-1}\left(X\right) + \theta _{k}  M_{k}\left(X\right)$}
                \EndFor
                \State \Return $M_{K}(X)$.

           \EndProcedure

        \end{algorithmic}

    \end{algorithm}

\section{Experimental Setup}
To demonstrate the effectiveness of the proposed CBPT for Twitter depression detection, we conduct experiments on two publicly accessible datasets: the Tsinghua Twitter Depression Dataset (TTDD) and the CLPsych 2015 Twitter Dataset (CLPsych2015). All experimental procedures have been approved by the Ethical Review body of University of Leicester. In this section, we describe the setup details of our evaluation.

\textbf{TTDD\footnote{http://depressiondetection.droppages.com/}}: The Twitter database was collected by Shen et al. \cite{shen2017depression} in 2017 for depression detection. The Twitter database has three parts: (1) \textbf{Depression Dataset D1}: The dataset was created based on the tweets collected between 2009 and 2016, where the users were labelled as depression if their anchor tweet satisfied the pattern "(I'm/I was/I am/I've been) diagnosed depression". (2) \textbf{Depression Dataset D2}: This dataset contains Twitter messages where users were labelled as non-depressed if they had never posted any tweets containing the character string "depress". (3) \textbf{Depression Dataset D3}: Shen et al. \cite{shen2017depression} constructed an unlabelled large dataset D3 for depression candidate. Based on the tweets shown in December 2016, this unlabelled depression candidate dataset was established where the user were recorded if their anchor tweet loosely contained the character string ``depress". There are 2558, 5304 and 58810 samples stored in D1, D2, D3, respectively. Each sample of these three datasets contains one-month post information of a Twitter user . In this paper, we employ the well labelled datasets D1 and D2 to evaluate our classification algorithm's performance and analyse the online behaviours of depression users.

\textbf{CLPsych 2015\footnote{http://www.cs.jhu.edu/~mdredze/clpsych-2015-shared-task-evaluation/}}: The dataset was established by John Hopkins University for a depression detection task in 2015 \cite{coppersmith-etal-2015-clpsych}. The dataset contains public Twitter users’ posts between 2008 and 2013 via the Twitter application programming interface (API). Similarly, possible mental disease sufferers are labeled as depression or post-traumatic stress disorder (PTSD) according to their self statement of diagnosis, such as "I was just diagnosed with depression or PTSD...". Furthermore, they conducted careful pre-preprocessing and anonymisation operations, such as filtering the users whose tweets are fewer than 25 and removing individual information. Finally, they manually examined and refined the annotation of each collected Twitter user’s logs by using a semi-supervised method. The processed dataset consists of 477 depressed users, 396 PTSD (an anxiety disorder caused by very stressful, frightening or distressing events) users and 873 control users. For each user, up to their most recent 3000 public tweets were included in the dataset.

\textbf{Implementation Details}: We implement the proposed CBDT and other benchmark experiments using the Scikit$\-$learn framework \cite{pedregosa2011scikit} and deploy all the experiments on a 8-core Intel Xeon skylake 2.6GHz CPU with 64GB RAM. The source code will be publicly accessible\footnote{https://github.com/BIPL-UoL/Cost-Boosting-Pruning-Trees-for-depression-detection-on-Twitter}.

\section{Experimental Results}
In this section, we present both quantitative and qualitative experimental results of different trials. We first conduct an ablation study of our method to show the impact of the pruning procedure and the cost-sensitive boosting scheme on the classification performance. We also compare our proposed Twitter depression detection framework with several state-of-the-art methods using the  aforementioned two Twitter datasets. Finally, we justify the signification factors for depression prediction by our model.
 
\begin{table*}[]
\centering
\scriptsize
\caption{Classification Results: [Mean Accuracy$/$F1 Score$\pm$Standard Deviation] by eight boosting classifiers for five public datasets. The Best results are shown in bold.}
\label{Tab:comparison_results}
\begin{tabular}{|c|l|l|l|l|l|l|l|l|l|l|}
\hline
\multicolumn{1}{|l|}{} & \multicolumn{2}{c|}{TTDD} & \multicolumn{2}{c|}{CLPsych 2015} & \multicolumn{2}{c|}{LSVT} & \multicolumn{2}{c|}{Statlog} & \multicolumn{2}{c|}{Glass} \\ \hline
Algorithm              & Accuracy    & F1-score    & Accuracy        & F1-score        & Accuracy    & F1-score    & Accuracy      & F1-score     & Accuracy    & F1-score    \\ \hline
Discrete Adaboost      & 86.48$\pm$0.93  & 84.88$\pm$1.02  & 64.76$\pm$2.02      & 61.28$\pm$2.48      &80.15$\pm$4.39      & 75.45$\pm$6.71     & 77.17$\pm$0.82    & 71.15$\pm$1.08   &  58.07$\pm$8.84            &            48.92$\pm$7.45  \\ \hline
Real Adaboost          &85.79$\pm$0.85             & 84.21$\pm$0.98           &   61.42$\pm$3.75              &  57.70$\pm$3.45               &81.72$\pm$3.30              & 78.05$\pm$5.09             &  70.34$\pm$4.29  & 62.54$\pm$4.26              &  40.64$\pm$11.68            &   29.72$\pm$19.01           \\ \hline
XGboost                &87.43$\pm$0.56             &  86.00$\pm$0.57           &     68.62$\pm$2.62             &  64.66$\pm$3.25                & 84.12$\pm$2.54          &  79.66$\pm$5.76            &     91.74$\pm$0.79  &   90.13$\pm$0.85            &    74.36$\pm$10.83          &    69.56$\pm$11.55         \\ \hline
LogitBoost             & 86.54$\pm$0.22            &  85.01$\pm$0.28           &   61.48$\pm$3.24              & 57.32$\pm$3.79                &80.09$\pm$6.80              & 76.00$\pm$5.65             &              90.33$\pm$0.63  & 88.23$\pm$0.59              & 75.27$\pm$6.74             &     71.84$\pm$8.98         \\ \hline
LightGBM               & 87.69$\pm$0.72           & 86.49$\pm$0.67            &  68.62$\pm$1.66               &  64.46$\pm$2.30               &85.75$\pm$3.87              &   79.90$\pm$10.72           &              \textbf{92.46$\pm$0.62}  & 90.90$\pm$0.59              &76.67$\pm$8.87              &  72.67$\pm$10.42            \\ \hline
KiGB                   & 87.73$\pm$0.68            & 86.29$\pm$0.68            &   67.06$\pm$2.05              &  62.79$\pm$2.27               &  81.69$\pm$5.53            &  77.76$\pm$4.83            &  91.40$\pm$0.70              &  89.71$\pm$0.59             & 77.13$\pm$8.94             &  67.87$\pm$12.84            \\ \hline
Adaboost+PT (Ours)     &  87.70$\pm$0.77          &   86.34$\pm$0.83          & 69.71$\pm$2.74                &   65.71$\pm$3.34              & \textbf{86.52$\pm$5.37}             &  \textbf{82.45$\pm$7.98}           &              87.13$\pm$1.05  &85.04$\pm$1.08               & \textbf{79.02$\pm$9.24 }            &            \textbf{72.70$\pm$9.06 } \\ \hline
CBPT (Ours)             &\textbf{88.39$\pm$0.60}     &  \textbf{86.90$\pm$0.62}       & \textbf{70.69$\pm$1.84}               &                 \textbf{66.54$\pm$2.42}  & 85.72$\pm$4.03           &  81.26$\pm$6.24           &     92.21$\pm$0.31          &   \textbf{91.20$\pm$0.38}            & 77.63$\pm$8.58             &   70.66$\pm$9.55           \\ \hline
\end{tabular}
\end{table*}

\begin{figure*}
     \centering
     \begin{subfigure}[b]{0.19\textwidth}
         \centering
         \includegraphics[width=1\textwidth]{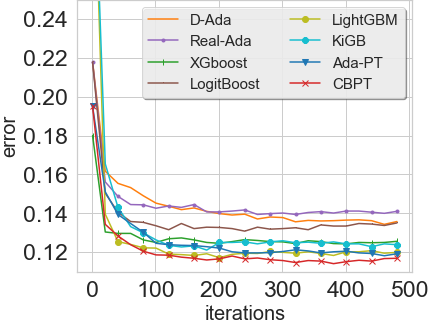}
         \caption{TTDD}
			\label{fig:ttdd_cr}
     \end{subfigure}
     \hfill
     \begin{subfigure}[b]{0.19\textwidth}
         \centering
         \includegraphics[width=1\textwidth]{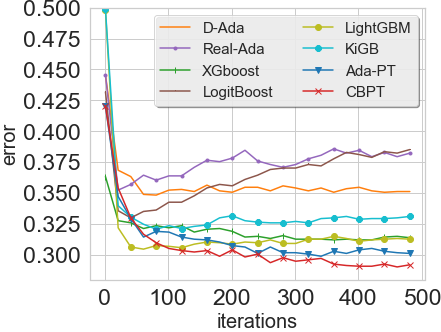}
         \caption{CLPsych 2015}
			\label{fig:clp_cr}
     \end{subfigure}
     \hfill
     \begin{subfigure}[b]{0.19\textwidth}
         \centering
         \includegraphics[width=1\textwidth]{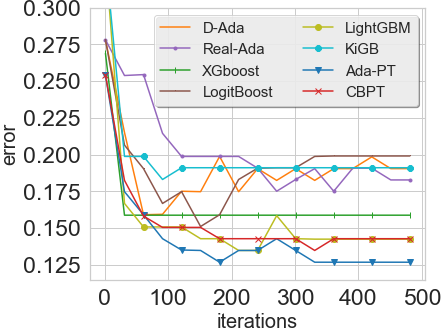}
         \caption{LSVT}
			\label{fig:lsvt_cr}
     \end{subfigure}
     \hfill
     \begin{subfigure}[b]{0.19\textwidth}
         \centering
         \includegraphics[width=1\textwidth]{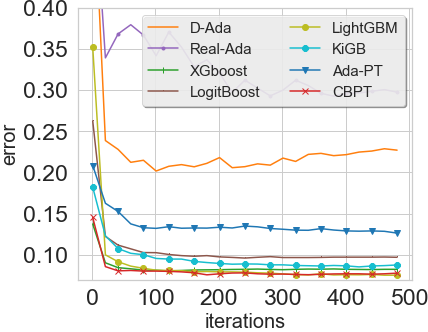}
         \caption{Statlog}
			\label{fig:statlog_cr}
     \end{subfigure}
     \hfill
     \begin{subfigure}[b]{0.19\textwidth}
         \centering
         \includegraphics[width=1\textwidth]{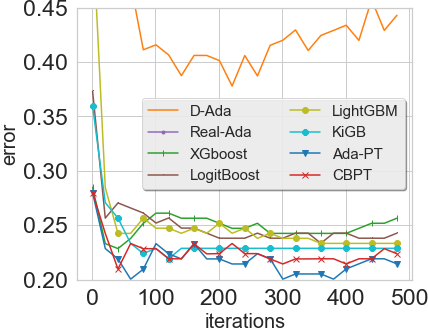}
         \caption{Glass}
			\label{fig:glass_cr}
     \end{subfigure}
        \caption{Convergence Rate: Testing error per iteration$/$tree.}
        \label{fig:convergence_curves}
\end{figure*}

\begin{figure*}
     \centering
     \begin{subfigure}[b]{0.19\textwidth}
         \centering
         \includegraphics[width=1\textwidth]{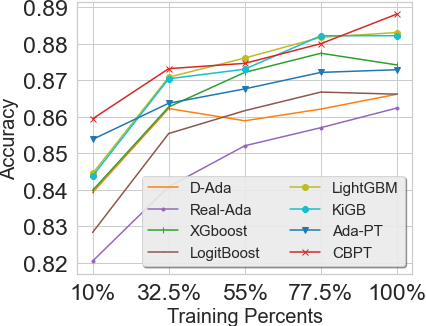}
         \caption{TTDD}
			\label{fig:ttdd_lc}
     \end{subfigure}
     \hfill
     \begin{subfigure}[b]{0.19\textwidth}
         \centering
         \includegraphics[width=1\textwidth]{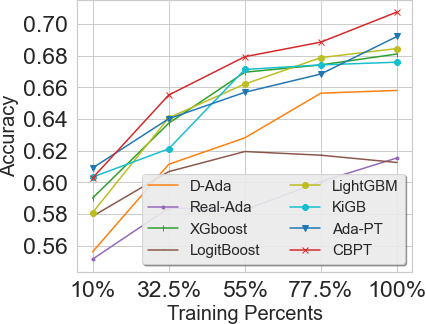}
         \caption{CLPsych 2015}
			\label{fig:clp_lc}
     \end{subfigure}
     \hfill
     \begin{subfigure}[b]{0.19\textwidth}
         \centering
         \includegraphics[width=1\textwidth]{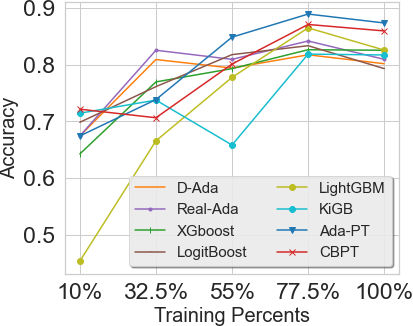}
         \caption{LSVT}
			\label{fig:lsvt_lc}
     \end{subfigure}
     \hfill
     \begin{subfigure}[b]{0.19\textwidth}
         \centering
         \includegraphics[width=1\textwidth]{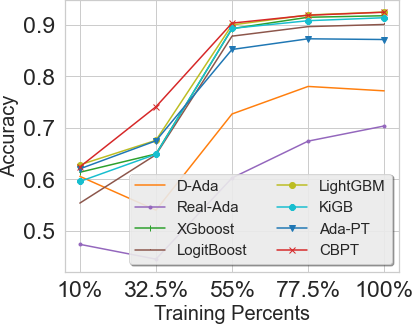}
         \caption{Statlog}
			\label{fig:statlog_lc}
     \end{subfigure}
     \hfill
     \begin{subfigure}[b]{0.19\textwidth}
         \centering
         \includegraphics[width=1\textwidth]{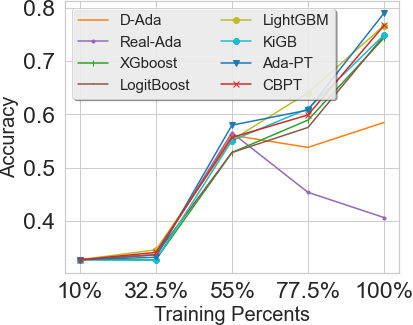}
         \caption{Glass}
			\label{fig:glass_lc}
     \end{subfigure}
        \caption{Learning curves for different training sets.}
        \label{fig:learning_curves}
\end{figure*}

\subsection{Ablation Studies}
In order to evaluate our proposed CBPT comprehensively, besides the two Twitter datasets, we also use three publicly accessible datasets (e.g. LSVT, Statlog, Glass) from the UCI machine learning repository \cite{asuncion2007uci} to examine our method's classification performance. We compare our method with Real Adaboost \cite{hastie2009multi}, XGboost \cite{chen2016xgboost}, LogitBoost \cite{li2012robust}, LightBoost \cite{ke2017lightgbm} and KiGB \cite{kokel2020unified}, which are state-of-the-art Boosting methods. We also investigate the performance of the standard Discrete Adaboost and combine the Discrete Adaboost structure with the pruning procedure (Adaboost+PT) as a comparison method to validate the effectiveness of our newly added components. We summarise the datasets' details in Table S5, Supplementary B.

For the performance comparison, we use Accuracy and F1-score as evaluation metrics. The UCI datasets have supplied feature vectors and the ground truth, so we use the same feature extraction procedure (aforementioned in Section 3) to extract features vectors from the two Twitter datasets. We use 5-fold cross-evaluation on the five datasets, where the training size is 75$\%$ and the test size is 25$\%$. To seek a fair  comparison, we have evaluated different settings of the hyperparameters for the compared methods and the best results on the test set are recorded. Some key hyperparameters include: (1) $Num \  leaves \in \left \{64,128,256 \right \}$, which control the size of each tree. (2) $Max \  depth \in \left \{5,10,15 \right \}$, which limit the maximum depth of each tree. (3) $Learning \  rate \in \left \{0.1,0.5,1 \right \}$, which determine the weight coefficient of each tree. (4) We fix the $tree \ number$ in all the classifiers to $500$ in order to obtain converging results. More details of the parameter setting are listed in Table S6-10, Supplementary B.

The results of classification on the five datasets are presented in Table \ref{Tab:comparison_results}. We observe that CBPT obtains the best performance in the two Twitter datasets and achieves 92.21$\%$ accuracy and a F1-score of 91.20$\%$ in the Statlog dataset. But in the LSVT and Glass datasets, the 'ablation' method Adaboost+PT results surpass CBPT by 1$\%$ and 2$\%$ separately. The reason is that the cost-sensitive boosting structure may be weak in the small-scale datasets. The Adaboost+PT outperforms the baseline Discrete Adaboost in the five datasets, confirming the effectiveness of our proposed pruning procedure. In general, the classification performance of CBPT for the five datasets is better than the other boosting methods except Adaboost+PT. To find out why this occurs, we undertake the following experiments.

Fig. \ref{fig:convergence_curves}(a)-(e) show the testing errors per iteration of the boosting classifiers for the five datasets. We observe that CBPT uses fewer trees to produce a comparable testing error in the TTDD, CLPsych 2015, and Statlog datasets. Comparing Adaboost+PT with CBPT, we witness the cost-sensitive boosting structure is effective to speed up the convergence of the algorithm in the TTDD, CLPsych 2015, and Statlog datasets. In the LSVT and Glass datasets, the cost-sensitive boosting structure is not helpful to improve the testing accuracy. As the LSVT and Glass datasets only have 128 and 214 samples respectively, we examine that in the cost-sensitive boosting structure, the newly added two penalty terms accelerate the weight updating and increase the variance in the small-scale datasets. To validate our assumption, we look at Fig. \ref{fig:learning_curves}(a)-(e). The accuracy of CBPT and Adaboost+PT increase as more training samples are added. In spite of being trained with small data, CBPT and Adaboost+PT still outperform the baseline Discrete Adaboost, which verifies the pruning procedure effectively improves the models' generalization ability. From Fig. \ref{fig:learning_curves}(a), (b) and (d), CBPT outperforms Adaboost+PT after having been trained with 32.5$\%$ or more training data. We summarise that in the case of sufficient training data, the proposed cost-sensitive boosting structure can improve the robustness of the model.

\subsection{Comparison with the SOTA Depression Detection Frameworks}
\begin{table}[]
\centering
\caption{Detection performance compared with the SOTA frameworks for the TTDD dataset. The best results are shown in bold.}
\label{Tab:ttdd_sota}
\begin{tabular}{|l|c|c|}
\hline
            & \multicolumn{2}{c|}{TTDD} \\ \hline
Method      & Accuracy    & F1-score    \\ \hline
Shen et al. \cite{shen2017depression} & 85$\%$           & 85$\%$         \\
Pedregosa et al. \cite{pedregosa2011scikit}               & 73$\%$              &            71$\%$   \\
Song et al. \cite{song2015multiple}             &  82$\%$           &    81$\%$         \\
 Rolet et al. \cite{rolet2016fast}            &  76$\%$            & 76$\%$             \\
  CBPT (Ours)              &    \textbf{88.39$\%$}             &    \textbf{86.90$\%$}         \\ \hline
\end{tabular}
\end{table}

\begin{table}[]
\centering
\caption{Detection performance compared with the SOTA frameworks for the CLPsych 2015 dataset. The best results are shown in bold. Columns: depression vs. control (DvC), depression vs. PTSD (DvP) and PTSD vs. control (PvC).}
\label{Tab:clp_sota}
\begin{tabular}{|l|c|c|l|}
\hline
                   & \multicolumn{3}{c|}{CLPsych 2015} \\ \hline
Method/Problem AUC & DvC       & DvP       & PvC       \\ \hline
Resnik et al.  \cite{resnik2015university}      &   0.860        & \textbf{0.841}        &          0.893 \\
 Preo{\c{t}}iuc-Pietro et al.       \cite{preoctiuc2015mental}    &  \textbf{0.862}        &            0.839 &   0.860        \\
 Pedersen et al. \cite{pedersen2015screening}              & 0.730          &          0.780 &   0.710         \\
   Coppersmith et al.          \cite{coppersmith-etal-2015-clpsych}    &        0.815 &   0.821        & 0.847          \\
    CBPT (Ours)                   &  0.840         &  0.812         &  \textbf{0.898} \\ \hline
\end{tabular}
\end{table}

In the above discussion, we have verified our proposed classifier CBPT outperforms the other SOTA boosting algorithms in the two Twitter depression detection datasets. We employ the same feature extraction procedure to extract features from the two Twitter datasets. We obtain 38 dimensional feature vectors from the TTDD dataset and 40 dimensional vectors from the CLPsych 2015 dataset (i.e. age and gender information are available so we extract the extra two features from the CLP dataset). The  two feature matrixes are used to train CBPT.

Tables \ref{Tab:ttdd_sota} and \ref{Tab:clp_sota} show the comparison results of depression detection. From Table \ref{Tab:ttdd_sota}, it is obvious that our framework achieves the best performance and surpasses the SOTA method of Shen et al. \cite{shen2017depression} by 3.39$\%$ on accuracy and 1.69$\%$ on F1-score. In the CLPsych 2015 leader-board, the detection performance is evaluated against three separate classification tasks, i.e. Depression vs. Control, Depression vs. PTSD and PTSD vs. Control. In Table \ref{Tab:clp_sota}, the CBPT results are competitive and better than the other methods in the PvC task. Another advantage of our framework is that the dimensionality of our extracted feature is far less than that of the other methods. For example, Resnik et al.  \cite{resnik2015university} employed a complicated Supervised LDA model to extract document vectors and combine these with large vocabularies (feature dimensionality is about 500). Preo{\c{t}}iuc-Pietro et al. \cite{preoctiuc2015mental} applied the unigram word features of 41687 dimensions to training their model. Our method only uses few features and achieves competitive performance for the CLPsych 2015 dataset. From the two comparison experiments, we can verify our proposed depression detection framework has satisfactory robustness on different datasets.

\begin{figure}
     \centering
     \begin{subfigure}[b]{0.23\textwidth}
         \centering
         \includegraphics[width=1\textwidth]{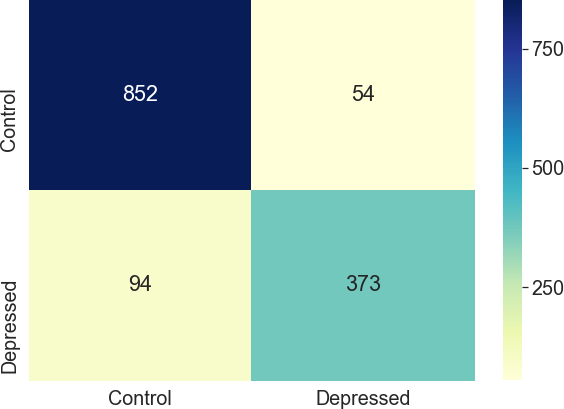}
         \caption{TTDD}
		
     \end{subfigure}
     \hfill
     \begin{subfigure}[b]{0.23\textwidth}
         \centering
         \includegraphics[width=1\textwidth]{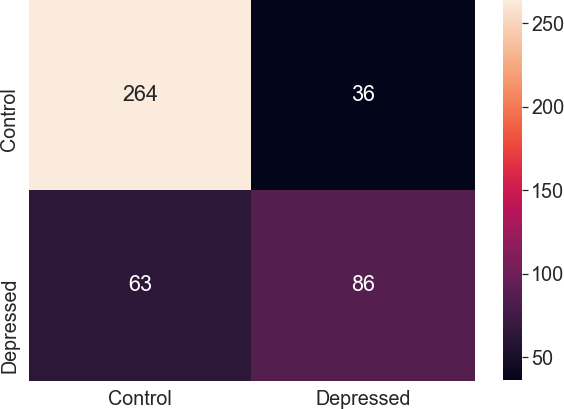}
         \caption{CLPsych 2015}
	
     \end{subfigure}

        \caption{Confusion Matrixes for the two depression detection datasets.}
        \label{fig:cm}
\end{figure}

\subsection{Explainable Depression Detection}

\begin{figure*}
     \centering
     \begin{subfigure}[b]{0.48\textwidth}
         \centering
         \includegraphics[width=1\textwidth]{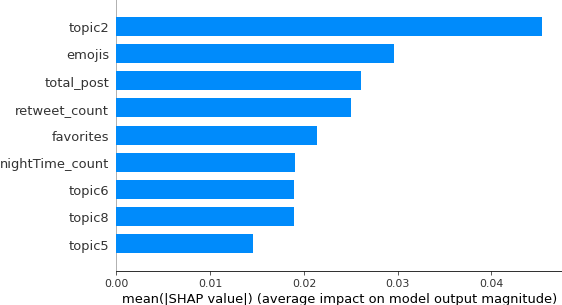}
         \caption{TTDD: Absolute feature importance.}
		
     \end{subfigure}
     \hfill
     \begin{subfigure}[b]{0.48\textwidth}
         \centering
         \includegraphics[width=1\textwidth]{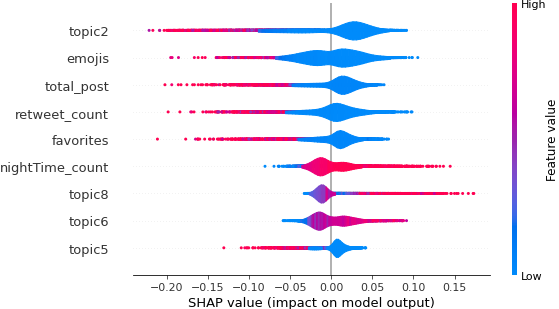}
         \caption{TTDD: Summary plot.}
		
     \end{subfigure}
     \hfill
     \begin{subfigure}[b]{0.48\textwidth}
         \centering
         \includegraphics[width=1\textwidth]{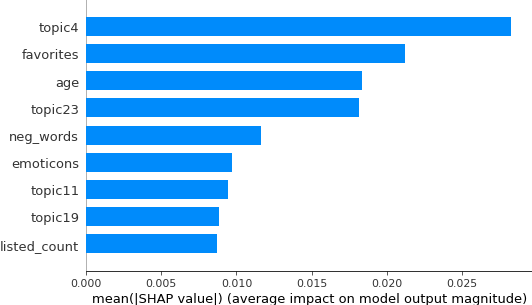}
         \caption{CLPsych 2015: Absolute feature importance.}
		
     \end{subfigure}
     \hfill
     \begin{subfigure}[b]{0.48\textwidth}
         \centering
         \includegraphics[width=1\textwidth]{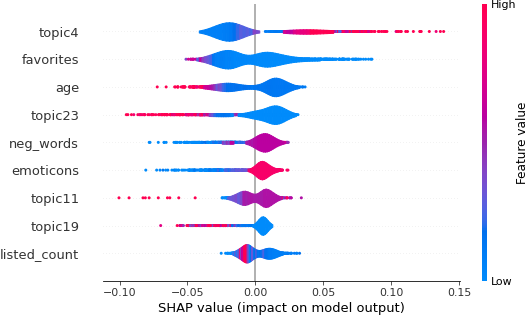}
         \caption{CLPsych 2015: Summary plot.}
			
     \end{subfigure}
     \hfill

        \caption{Top 9 significant features for depression detection. (a) and (c):  Average feature importance. (b) and (d): Summary Plots. Each point is a Shapley Value $\phi_{X_{i}^{v}}(f)$ corresponding to a feature and an instance. Overlapping points are jittered on the y-axis direction.}
        \label{fig:feature_importances}
\end{figure*}

\begin{figure}
     \centering
     \begin{subfigure}[b]{0.5\textwidth}
         \centering
         \includegraphics[width=1\textwidth]{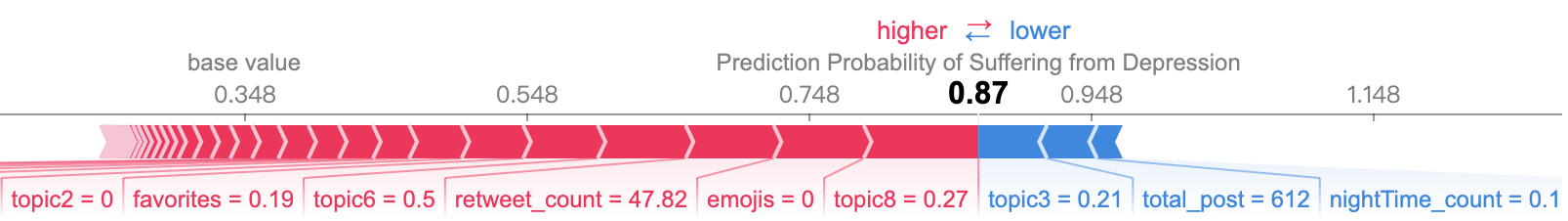}
         \caption{A Depressed User.}
			\label{fig:ttdd_predict}
     \end{subfigure}

          \begin{subfigure}[b]{0.5\textwidth}
         \centering
         \includegraphics[width=1\textwidth]{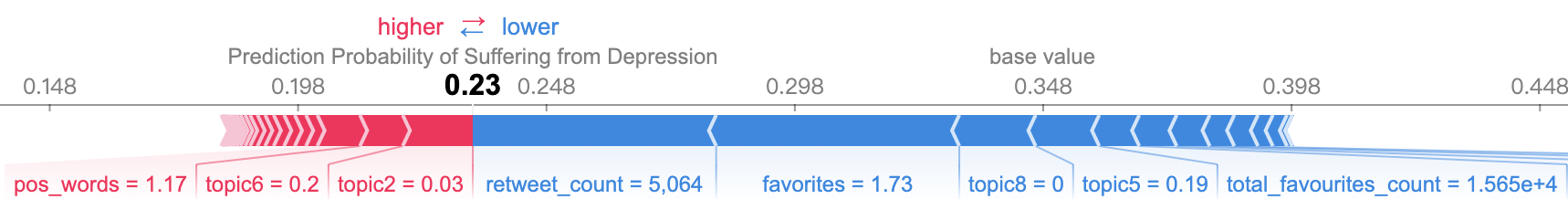}
         \caption{A Control User.}
			\label{fig:clp_predict}
     \end{subfigure}


        \caption{Additive force plots for two Twitter users. The bold text is the predicted depression probability. Each arrow (red or blue) is a single feature of the instance and the arrow length represents the feature's Shapley Value.}
        \label{fig:sing_predict}
\end{figure}

Previous research studies\cite{shen2017depression,nadeem2016identifying,jamil2017monitoring} have widely analysed online behaviours of depressed users through examining features' distributions or mean values and variances. But they have not explored which specific factors contribute to depression detection. Tree Shapley Additive Explanation (TreeSHAP) \cite{lundberg2020local} is a game approach to explain the output of decision trees based models. The goal of TreeSHAP is to explain the prediction of any instance by measuring the contribution of each feature to the prediction. TreeSHAP treats Shapley Values \cite{shapley1951notes} as the features' contributions and uses all the advantages of Shapley Values: (1) TreeSHAP has a solid theoretical foundation in the game theory. (2) The prediction is fairly distributed over the features' values. (3) TreeSHAP gives contrastive explanations that compares the prediction with the model's expectation\cite{molnar2020interpretable}. Hence, we integrate our framework with TreeSHAP to comprehensively investigate the influencing factors for the prediction results. We use the subset CvD of the CLPsych 2015 and the TTDD dataset for evaluating the depression risk factors, and other results (e.g. DvP, PvC subsets) are shown in Supplementary C. Besides, we list the related formulas of TreeSHAP in Section A, Supplementary D  and we give an example for the calculation of the Shapley Values via decision trees in Section B, Supplementary D. 

In Section A, Supplementary D, we describe that $\phi_{X_{i}^{v}}(f)$ represents the contribution of feature $X_{i}^{v}$ to the classifier's prediction for instance $X_{i}$. In our depression detection datasets, we aim to explore the influencing factors for the predicted depression risk of Twitter users, so the value of $\phi_{X_{i}^{v}}(f)$ represents how much the predicted depression probability for instance $X_{i}$ has been affected by feature $X_{i}^{v}$. 

\begin{figure*}
     \centering
     \begin{subfigure}[b]{0.32\textwidth}
         \centering
         \includegraphics[width=1\textwidth]{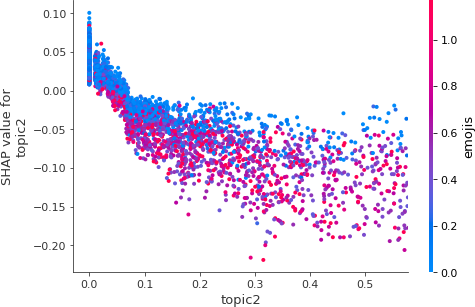}
         \caption{topic2 interacts with emojis.}
		
     \end{subfigure}
     \hfill
     \begin{subfigure}[b]{0.32\textwidth}
         \centering
         \includegraphics[width=1\textwidth]{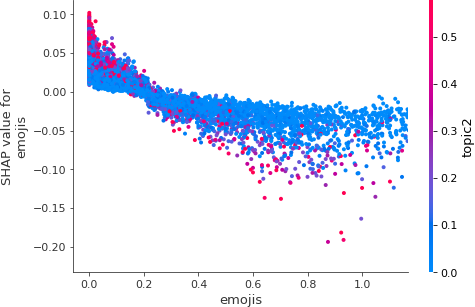}
         \caption{emojis interacts with topic2.}
		
     \end{subfigure}
     \hfill
     \begin{subfigure}[b]{0.32\textwidth}
         \centering
         \includegraphics[width=1\textwidth]{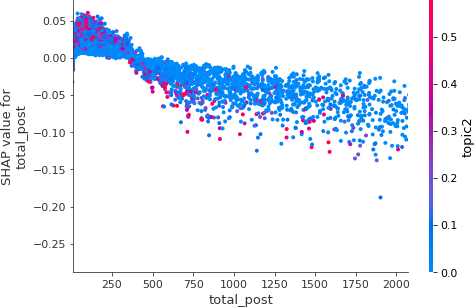}
         \caption{total$\_$post interacts with topic2.}
		
     \end{subfigure}
     \hfill
     \begin{subfigure}[b]{0.32\textwidth}
         \centering
         \includegraphics[width=1\textwidth]{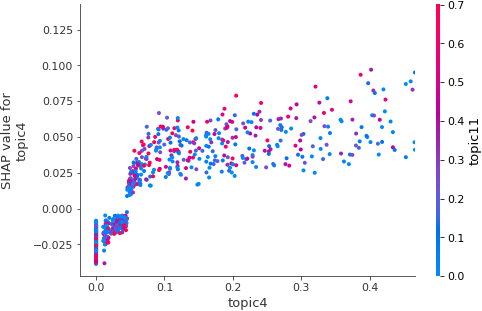}
         \caption{topic4 interacts with topic11.}
			
     \end{subfigure}
     \hfill
     \begin{subfigure}[b]{0.32\textwidth}
         \centering
         \includegraphics[width=1\textwidth]{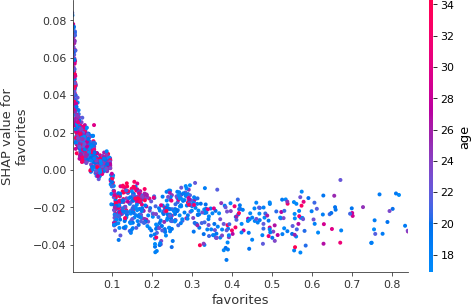}
         \caption{favourites interacts with age.}
			
     \end{subfigure}
     \hfill
     \begin{subfigure}[b]{0.32\textwidth}
         \centering
         \includegraphics[width=1\textwidth]{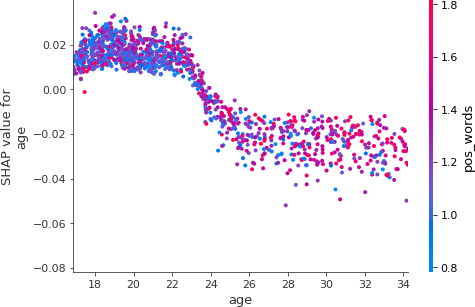}
         \caption{age interacts with pos$\_$words.}
			
     \end{subfigure}
     \hfill

        \caption{Feature dependence Plot: The point’s position on the X axis represents the target feature value, and the values on the Y axis are the Shapley Values for the target feature. Color bar is the value range of the interactive feature.  (a)-(c) are the dependence plots of the TTDD dataset. (d)-(f) are the dependence plots of the CLPsych 2015 dataset.Please note that the x-axis and color bar range have been trimmed to the 5th and 95th percentile of the data in order to avoid the x-axis or color bar being too board because of the outliers.}
        \label{fig:dependence_plots}
\end{figure*}

Fig. \ref{fig:cm} shows two confusion matrices of the prediction results of CBPT over the two depression detection datasets. From these figures, we know how many depressed or control users has been classified. Here, we use feature importance to analyse which feature significantly affects global depression detection. Feature importance is computed by $I^{v} = \frac{\sum_{i=1}^{N}\left |\phi_{X_{i}^{v}}(f)  \right |}{N}$ (N is the number of data instances). Fig. \ref{fig:feature_importances}(a) and (c) show top 9 significant features for depression detection in the two Twitter datasets. In these two figures, features with large absolute Shapley Values are important. For example, topic2 stands in the most critical position in Fig. \ref{fig:feature_importances}(a) and topic2 changes the predicted depression probability by 4\% on average for all the instances. Although the feature importance plot is useful, there is no more information beyond the importance. For more information, we use summary plots (Fig. \ref{fig:feature_importances}(b) and (d)) to further analyse the significant features. In the summary plots, each point is a Shapley Value $\phi_{X_{i}^{v}}(f)$ corresponding to a feature and an instance. Overlapping points are jittered on the y-axis direction so each row is the distribution of Shapley Values. In Fig. \ref{fig:feature_importances}(b), topic2 with a high feature value (red points) stands for decreasing depression risk and a low value of topic2 (blue points) refers to increasing depression risk. In Table S13-S14, Supplementary C,  we show the top 10 words that are the most likely to occur in each LDA topic. Topic2 includes words such as 'trump', 'obama', 'russia' that infers topic2 may be related to 'politics'. The feature value of topic2 is the occurrence probability of topic2 in the posting texts. If a user posts many tweets on the theme of politics, his/her predicted depression risk will be decreased. Similarly, if a user posts many tweets with emojis that receive many retweets, the user is less likely to be depressed. Depressed users seem to lack of communication with others that depressed users are more likely to post tweets during midnight and their posted tweets are barely retweeted or favoured by other users. And it is an interesting phenomenon that the posting texts of depressed users may involve the content of 'film' (topic8) or 'policy' (topic6) but without 'band' (topic5) information. In Fig. \ref{fig:feature_importances}(d), the most important feature topic4 is related to the theme of 'mental health' (shown in Table S14, Supplementary C). In the CLPsych 2015 dataset, depressed users are more likely to undertake the following behaviours: (1) Their posted tweets are related to the topics of 'mental health' (topic5) or 'news' (topic11) and include many emoticons and negative words. (2) They are young and they do not take many Twitter activities. (3) Their posted tweets may not be favoured by others and their tweets' content is not related to 'friend' (topic19)  and 'autism'(topic23).

Then, we use the additive force plots to explain why a user is predicted as depressed or control. Using two instances from the TTDD dataset, Fig. \ref{fig:sing_predict}(a) is the prediction visualization of a depressed user. In Fig. \ref{fig:sing_predict}(a), the bold text 87\% is the predicted depression probability and the base value 34.8\% is the classifier's expectation $\phi_{0}(f)$ referring to Eq. (2), Supplementary D. Features pushing the prediction higher are shown in red, while those pushing the prediction lower are shown in blue. For example, the emoji number of this user is 0 which is lower than the average value 0.34 (shown in Table S2, Supplementary A) and it contributes 8\% probability to the depression prediction. The total$\_$post feature (= 612 that is larger than 457.31)\footnote{612 is the feature value of total$\_$post and 457.31 is the average value of this feature (shown in Table S2, Supplementary A).} reduces the predicted risk about 4\%. This supports our finding in Fig. \ref{fig:feature_importances}(a)-(b) that few emojis lead to higher predicted depression risk and posting many tweets leads to less risk. Similarly, for a control user shown in Fig. \ref{fig:sing_predict}(b), this user's posting content may not be relevant to 'politics' (topic2=0.03 that is less than 0.11) that increases the depression risk by 1\%. This users' tweets are frequently retweeted by others (retweet$\_$count=5064 that is over 1843.14) and this behaviour decreases the user's predicted depression risk. The predicted depression probability for the control user drops from the base probability 34.8\% to 23\%.

Finally, we use the dependence plots to show the detailed interpretation of the features' impacts. Fig. \ref{fig:dependence_plots} includes 6 dependence plots for the most important three features with their most interactive features in the two datasets. The interactive feature can be selected arbitrarily and we decide the most interactive features depending on Eq. (6), Supplementary D. This equation calculates the correlation coefficient between the Shapley Values of the target feature and the values of the other features. In Fig. \ref{fig:dependence_plots}(a),  the predicted depression risk decreases with the increasing of the values of topic2 and emojis. This suggests that posting tweets on the theme of politics with many emojis leads to lower predicted depression risks and vice versa. In Fig. \ref{fig:dependence_plots}(b), topic2 is also the most interactive feature of emojis. This figure shows a similar trend to Fig. \ref{fig:dependence_plots}(a). In Fig. \ref{fig:dependence_plots}(c), the predicted depression risk shows a decreasing value at total$\_$post=400. This suggests that control users are more likely to post many tweets and share politics news than the depressed users. Similarly, from Fig. \ref{fig:dependence_plots}(d)-(f), we observe that posting tweets about 'mental health' (topic4) and 'news' (topic11) is proportionally related to the predicted depression risk. Depressed users' tweets are hard to receive favourites from others. The predicted depression risk for the Twitter users is decreasing at age=23 and using the positive or negative words will change the depression risk of the users. By the above feature dependence analysis, we have shown the influences of the feature interactions on the classifier's predicted depression probability and revealed the difference of the online behaviours between the depressed and control users.

\section{Conclusion}
In this paper, we have made an attempt to automatically identify potential Twitter depressed users. As we have known, most of the established works mainly focused on exploring new features of depression behaviours whilst ignoring the fitness of the classification models. Considering the complexity of Twitter data, in order to improve the robustness of the decision tree based estimator, we proposed a novel resampling weighted pruning algorithm which dynamically determines optimal depths/layers and leaves of a tree model. Taking into account the "hardness" of different misclassified samples, we also proposed a cost-sensitive boosting structure to hierarchically update the instances' weights in the pruned trees. We combined the proposed pruning process with the novel cost-sensitive boosting structure within an ensemble framework, namely Cost-sensitive Boosting Pruning Trees (CBPT) to classify control and depressed users.

CBPT outperformed the other depression detection frameworks in the two Twitter datasets. In the meantime, we conducted the convergence analysis of our proposed CBPT through comprehensive experiments. Moreover, we utilised three UCI datasets to evaluate the classification ability of our method quantitatively, which shows our method performs better than the other SOTA boosting algorithm. We then integrated CBPT with TreeSHAP in order to explain the predicted depression risks of Twitter users by investigating the contribution of each feature to the prediction. We used three different types of figures, i.e. additive force plot, summary and dependence plots, to explain the contributions of individual features to the predicted depression risks.

Taking a close look at the above experimental results, we found that the features extracted from the tweet content were really important for depression prediction. Features including LDA topics, negative/positive words and emojis play a key role in online depression risk detection. In the future, we will develop a robust topic model methodology to summarise posting text content of depressed users with clearly explainable topics. We will also attempt to mine similar information over other social networks, e.g. Facebook, Instagram, and Tumblr, for sentiment analysis.
\section*{Acknowledgements}
The work of Huiyu Zhou is supported in part by the Royal Society-Newton Advanced Fellowship under Grant NA160342.
\bibliographystyle{IEEEtranTIE}
\bibliography{bibliography}
\vspace{-1.5 cm}
\begin{IEEEbiography}[{\includegraphics[width=1in,height=1.1in,clip,keepaspectratio]{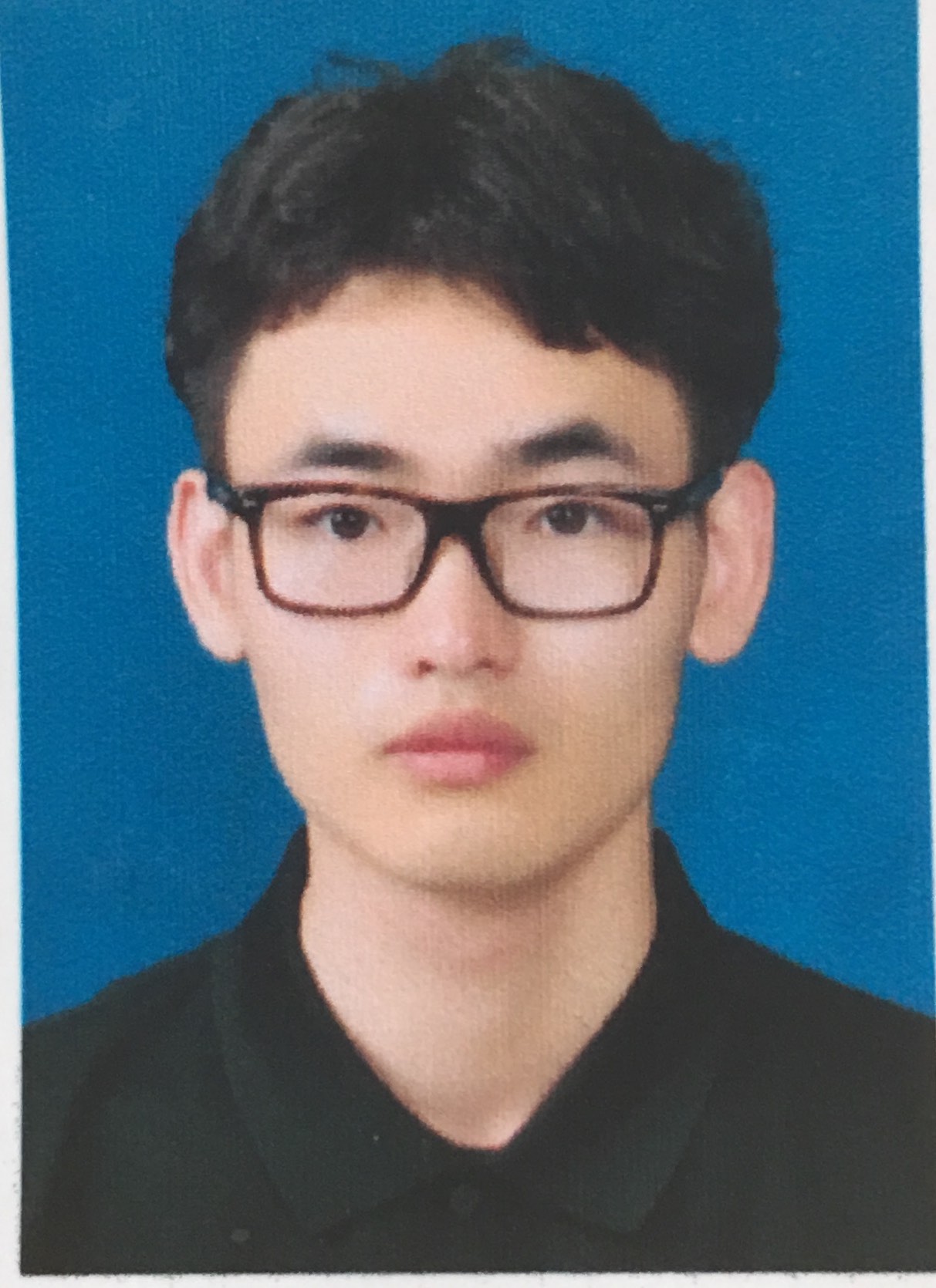}}]
{Lei Tong} is currently pursuing the Ph.D. degree with the School of Informatics, University of Leicester, Leicester, U.K.
His research interests include computer vision, social network analysis and data mining.
\end{IEEEbiography}
\vspace{-1.5 cm}
\begin{IEEEbiography}
[{\includegraphics[width=1in,height=1.1in,clip,keepaspectratio]{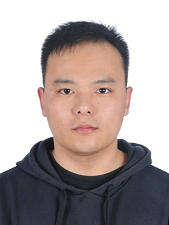}}]
{Zhihua Liu} is currently pursuing the Ph.D. degree with the School of Informatics, University of Leicester, Leicester, U.K. His research interests include machine learning, deep learning and computer vision.
\end{IEEEbiography}
\vspace{-1 cm}
\begin{IEEEbiography}
[{\includegraphics[width=1in,height=1.1in,clip,keepaspectratio]{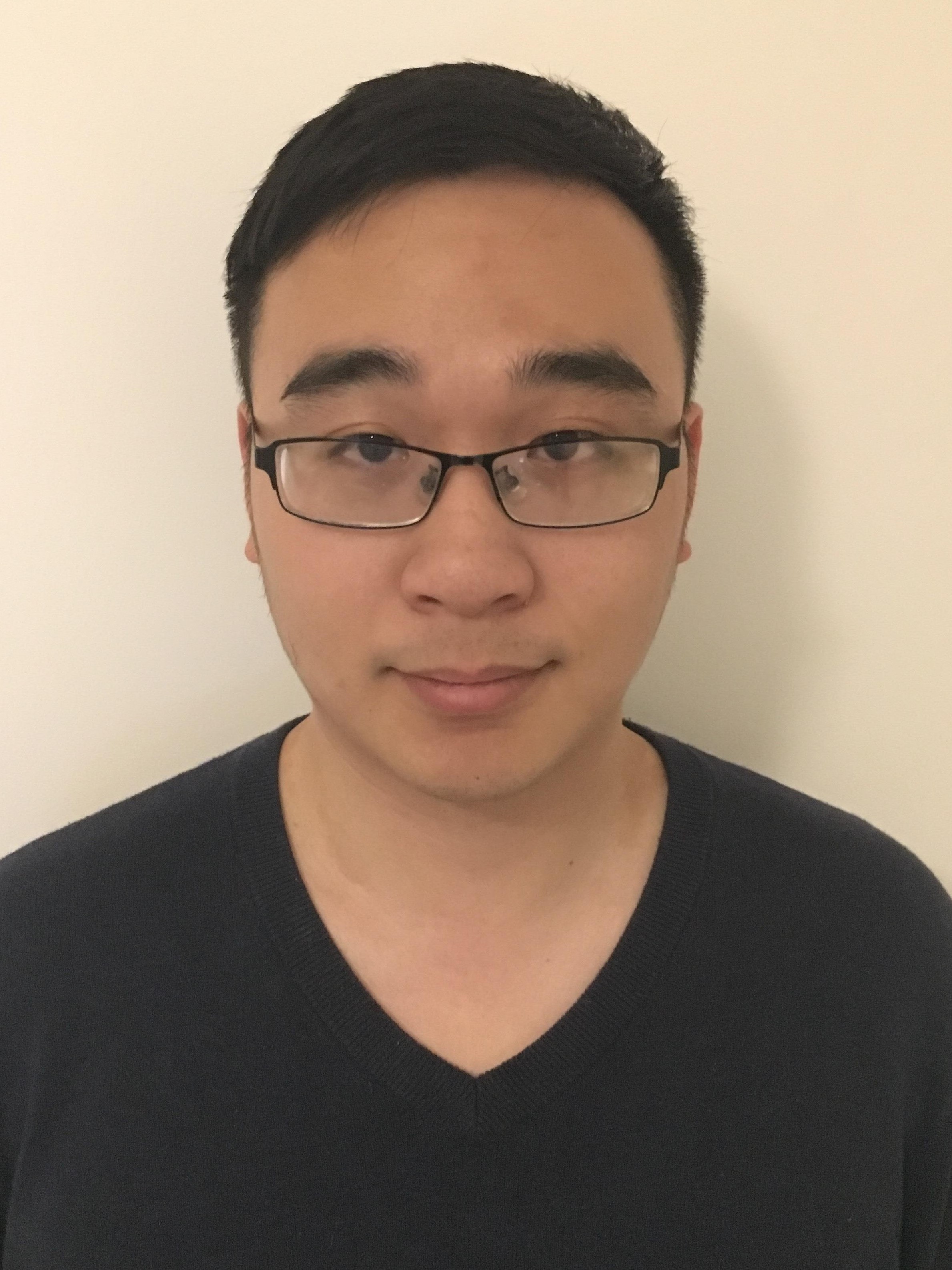}}]
{Zheheng Jiang} has been awarded his Ph.D. degree in Computer Science from University of Leicester, Leicester, U.K. He is currently the Senior Research Associate at the Computing and Communications, Lancaster University, Lancaster, U.K. His current research interests include machine learning for vision, object detection and recognition, video analysis and event recognition.
\end{IEEEbiography}
\vspace{-1 cm}
\begin{IEEEbiography} 
[{\includegraphics[width=1in,height=1.1in,clip,keepaspectratio]{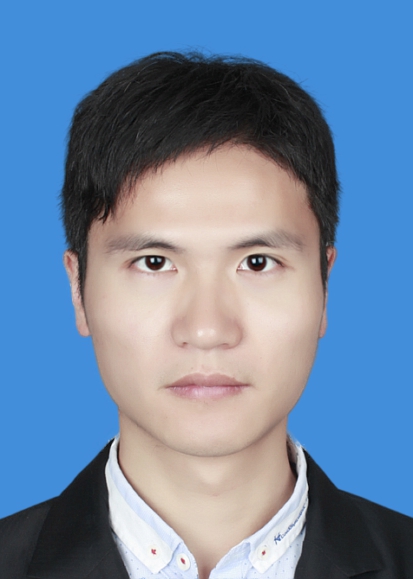}}]
{Feixiang Zhou}
is currently pursuing the Ph.D. degree with the School of Informatics, University of Leicester, Leicester, U.K. His current research interests include Computer Vision, Machine Learning and their applications on video understanding.
\end{IEEEbiography}
\vspace{-1 cm}
\begin{IEEEbiography}
[{\includegraphics[width=1in,height=1.1in,clip,keepaspectratio]{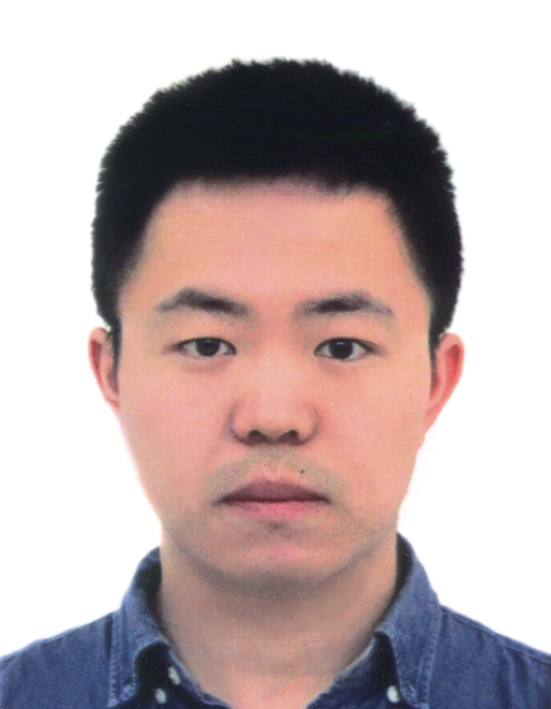}}]
{Long Chen}
 is currently pursuing the PhD degree with the School of Informatics, University of Leicester, U.K. His research interests are in the areas of Computer Vision and Machine Learning.
\end{IEEEbiography}
\vspace{-1cm}
\begin{IEEEbiography}
[{\includegraphics[width=1in,height=1.1in,clip,keepaspectratio]{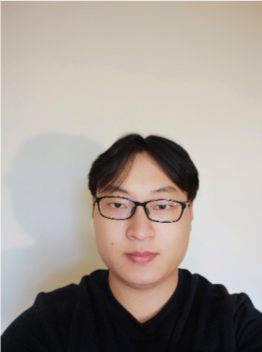}}]
{Jialin Lyu} is an MPhil student at the School of Informatics, University of Leicester. His current project is Automated Classification of Alzheimer's Disease and Mild Cognitive Impairment Using Deep Neural Networks. His research interests include machine learning and medical image analysis.
\end{IEEEbiography}
\vspace{-1 cm}
\begin{IEEEbiography}[{\includegraphics[width=1in,height=1.1in,clip,keepaspectratio]{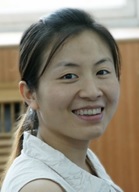}}]{Xiangrong Zhang}
received the B.S. and M.S. degrees from the School of Computer Science, Xidian University, Xi'an, China, in 1999 and 2003, respectively, and the Ph.D. degree from the School of Electronic Engineering, Xidian University, in 2006. Currently, she is a professor in the Key Laboratory of Intelligent Perception and Image Understanding of the Ministry of Education, Xidian University, China. She has been a visiting scientist in Computer Science and Artificial Intelligence Laboratory, MIT between Jan. 2015 and March 2016. Her research interests include pattern recognition, machine learning, and remote sensing image analysis and understanding.
\end{IEEEbiography}
\vspace{-1 cm}
\begin{IEEEbiography}[{\includegraphics[width=1in,height=1.1in,clip,keepaspectratio]{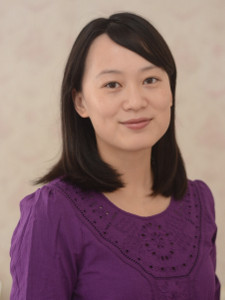}}]{Qianni Zhang}
received the Ph.D. degree from Queen Mary University of London, U.K., in 2007. She is currently a Senior Lecturer (Associate Professor) at the School of Electronic Engineering and Computer Science, Queen Mary University of London. She has authored over 50 technical papers and book chapters, and has actively contributed to several European funded research projects. Her research interests include multimedia processing, semantic inference and reasoning, machine learning, image understanding, 3D reconstruction, and immersive environments. She has served as a Guest Editor of a special issue in Journal of Multimedia and a Reviewer of journals including the IEEE Transactions on CSVT, Image Processing, Multimedia, Sensors, Signal Processing: Image Communication, and various conferences and workshops including the IEEE ICIP, ICASSP, ICME and ACM Multimedia. She has served as an organiser, a session chair, or a member of the technical program committee of several international conferences, workshops, or special sessions.
\end{IEEEbiography}
\vspace{-1 cm}
\begin{IEEEbiography}[{\includegraphics[width=1in,height=1.1in,clip,keepaspectratio]{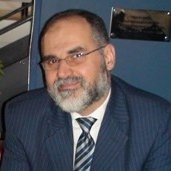}}]{Abdul Sadka}
is the Director of the Institute of Digital Futures, Leader of the Media Communications Research Group and the former Head of the Department of Electronic and Computer Engineering (Oct 06 - Jul 12) at Brunel University London. He has a 25-years research leadership focused  academic career underlined by a commitment to business engagement in academia. He has 200+ publications in refereed journals and conferences, 3 patents and a seminal textbook entitled ”Compressed Video Communications” published by J. Wiley in 2002. He has thus far managed to attract circa £13m worth of research grants and contracts and supervised nearly 50 Research Assistants and PhD students to full completion. He frequently serves on influential advisory boards and international evaluation panels and provides expert consultancy services to the Telecom/ICT industry as well as corporate Law firms in the area of 2D/3D video compression and multimedia processing. He serves on the Steering Board of the NEM European technology platform which helps shape the strategic research agenda of the European Commission Horizon research programmes. He is a Chartered Engineer (CEng), Fellow of HEA, IET and BCS.
\end{IEEEbiography}
\vspace{-1 cm}
\begin{IEEEbiography}
[{\includegraphics[width=1in,height=1.1in,clip,keepaspectratio]{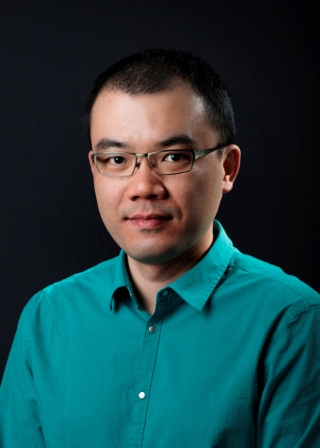}}]{Yinhai Wang}
received the B.Eng. degree in computer software from Jinan University, Guangzhou, China, in 2001, the M.Sc. degree in software engineering from Napier University, Edinburgh, U.K., in 2002, and the Ph.D. degree in electronics, electrical engineering, and computer science from Queen’s University, Belfast, U.K., in 2008. He is currently an Associate Director at Data Science $\&$ Quantitative Biology, AstraZeneca (Cambridge, UK) focusing on the use of image analysis and data sciences in biological applications. 
\end{IEEEbiography}
\vspace{-1 cm}
\begin{IEEEbiography}[{\includegraphics[width=1in,height=1.1in,clip,keepaspectratio]{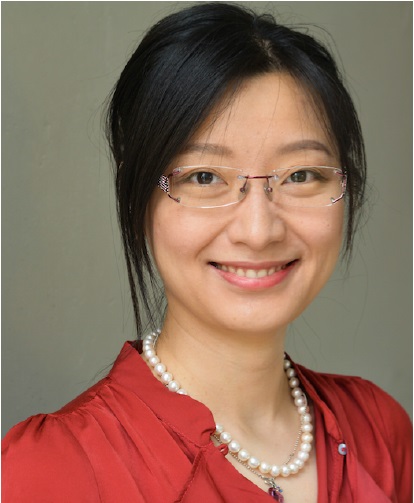}}]{Ling Li}
is the Director of Internationalisation at the School of Computing and also the founding coordinator of Laboratory of Brain $\mid$ Cognition $\mid$ Computing (BC2 Lab) of the school responsible for coordinating multidisciplinary research between Computing, Sports and local NHS hospitals. She had six-year research experience at Imperial College London with a focus to understand body sensor data (EEG, EMG, ECG, eAR-sensor, and etc.). She participated in large scale projects. She also involved in projects from government and industry (i.e. Samsung GRO award). She now serves at the editorial board of Brain Informatics and the secretary of IEEE Computing Society in UK and Ireland.
\end{IEEEbiography}
\vspace{-1 cm}
\begin{IEEEbiography}[{\includegraphics[width=1in,height=1.1in,clip,keepaspectratio]{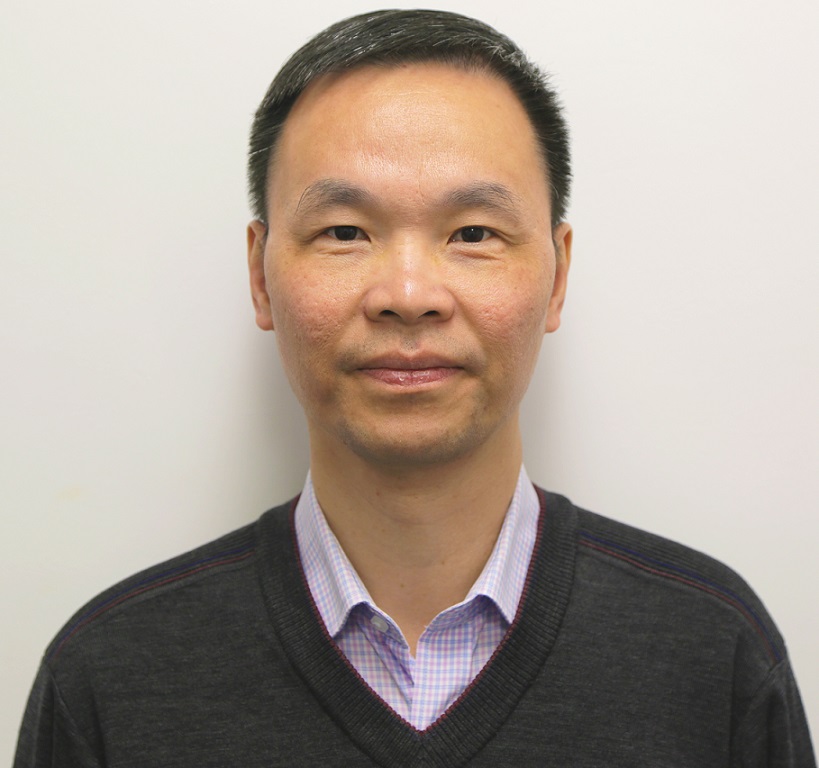}}]{Huiyu Zhou}
received a Bachelor of Engineering degree in Radio Technology from Huazhong University of Science and Technology of China, and a Master of Science degree in Biomedical Engineering from University of Dundee of United Kingdom, respectively. He was awarded a Doctor of Philosophy degree in Computer Vision from Heriot-Watt University, Edinburgh, United Kingdom. Dr. Zhou currently is a full Professor at School of Informatics, University of Leicester, United Kingdom. He has published over 350 peer-reviewed papers in the field. His research work has been or is being supported by UK EPSRC, ESRC, AHRC, MRC, EU, Royal Society, Leverhulme Trust, Puffin Trust, Invest NI and industry.
\end{IEEEbiography}

\clearpage
\onecolumn

\setcounter{table}{0}
\setcounter{figure}{0}
\setcounter{equation}{0}
\setcounter{page}{1}
\renewcommand\thefigure{S\arabic{figure}}
\renewcommand\thetable{S\arabic{table}}

\section{Supplementary A: Related Works and Extracted Features}
  \label{Appendix_A}
    Table \ref{Tab:feature_table} shows the features dimensionality information.
    
Table \ref{Tab:feature_value1} shows the maximum, minimum, mean and median values of feature variables for the two Twitter depression datasets.

Table \ref{Tab:feature_value2} shows the maximum, minimum, mean and median values of feature variables of individual classes for the two Twitter depression datasets.

  Table \ref{Tab:literature_compare} summarises the methods of depression detection introduced in Section 2 of the main manuscript.

\begin{table}[htbp]
\centering
\caption{Feature dimensionality.}
\label{Tab:feature_table}
\begin{tabular}{|c|l|c|}
\hline
\multicolumn{1}{|l|}{Feature Category}             & Feature Name      & \multicolumn{1}{l|}{Dimensions} \\ \hline
\multirow{4}{*}{{User Profile Features}}       & total favourites  & 1                               \\ \cline{2-3}
                                                   & friends           & 1                               \\ \cline{2-3}
                                                   & followers         & 1                               \\ \cline{2-3}
                                                   & listed count      & 1                               \\ \hline
\multirow{5}{*}{{Social Interaction Features}} & favourites count  & 1                               \\ \cline{2-3}
                                                   & retweet count     & 1                               \\ \cline{2-3}
                                                   & mention count     & 1                               \\ \cline{2-3}
                                                   & posting number    & 1                               \\ \cline{2-3}
                                                   & time distribution & 1                               \\ \hline
\multirow{5}{*}{{Linguistic Features}}         & negative words    & 1                               \\ \cline{2-3}
                                                   & positive words    & 1                               \\ \cline{2-3}
                                                   & emoji number      & 1                               \\ \cline{2-3}
                                                   & emoticons         & 1                               \\ \cline{2-3}
                                                   & LDA topics        & 25                              \\ \hline
\multicolumn{1}{|l|}{(only used in CLPsych 2015)}  & age               & 1                               \\ \hline
\multicolumn{1}{|l|}{(only used in CLPsych 2015)}  & gender            & 1                               \\ \hline
\multicolumn{2}{|c|}{Total dimensions}                                 & \begin{tabular}[c]{@{}l@{}}38 dimensional features for the TTDD dataset or\\40
dimensional features for the CLPsych2015 dataset                           \end{tabular}
 \\ \hline
\end{tabular}
\end{table}

\begin{table}[htbp]
\centering
\caption{Feature Statistics for the two Twitter depression datasets. The samples of the ptsd class in the CLPsych2015 datasets are removed as we only use the CvD subset here.}
\label{Tab:feature_value1}
\begin{tabular}{|c|cccc|cccc|}
\hline
\multirow{2}{*}{\textbf{Features}}       & \multicolumn{4}{c|}{\textbf{TTDD}}                                                                 & \multicolumn{4}{c|}{\textbf{CLPsych 2015}}                                                        \\ \cline{2-9} 
                                & \multicolumn{1}{c|}{max} & \multicolumn{1}{c|}{min} & \multicolumn{1}{c|}{mean} & median  & \multicolumn{1}{c|}{max} & \multicolumn{1}{c|}{min} & \multicolumn{1}{c|}{mean} & median  \\ \hline
total   favourites              & 356830.00                & 0.00                     & 6660.35                   & 1535.50 & 226236.00                & 0.00                     & 2121.72                   & 321.50  \\ \cline{1-1}
friends                         & 61572.00                 & 0.00                     & 678.67                    & 328.00  & 72296.00                 & 0.00                     & 847.05                    & 324.00  \\ \cline{1-1}
followers                       & 102069.00                & 0.00                     & 1039.58                   & 351.00  & 234328.00                & 0.00                     & 1742.32                   & 321.50  \\ \cline{1-1}
listed count                    & 1722.00                  & 0.00                     & 17.95                     & 4.00    & 2156.00                  & 0.00                     & 19.38                     & 1.00    \\ \cline{1-1}
favourites   count              & 1449.55                  & 0.00                     & 0.90                      & 0.24    & 127.77                   & 0.00                     & 0.38                      & 0.09    \\ \cline{1-1}
retweet count                   & 169788.00                & 0.00                     & 1843.14                   & 327.46  & 40743.00                 & 0.00                     & 618.18                    & 81.36   \\ \cline{1-1}
mention count                   & 5.52                     & 0.00                     & 0.69                      & 0.67    & 2.68                     & 0.00                     & 0.69                      & 0.64    \\ \cline{1-1}
posting number                  & 2972.00                  & 6.00                     & 457.31                    & 179.00  & 3000.00                  & 15.00                    & 2104.15                   & 2668.00 \\ \cline{1-1}
time   distribution             & 1.00                     & 0.00                     & 0.31                      & 0.32    & 1.00                     & 0.00                     & 0.35                      & 0.38    \\ \cline{1-1}
negative words                  & 3.68                     & 0.00                     & 0.69                      & 0.65    & 2.01                     & 0.00                     & 0.66                      & 0.64    \\ \cline{1-1}
positive words                  & 5.50                     & 0.00                     & 1.31                      & 1.27    & 3.05                     & 0.00                     & 1.24                      & 1.22    \\ \cline{1-1}
emoji number                    & 9.31                     & 0.00                     & 0.34                      & 0.18    & 0.00                     & 0.00                     & 0.00                      & 0.00    \\ \cline{1-1}
emoticons                       & 3.00                     & 0.00                     & 0.25                      & 0.23    & 4.00                     & 1.11                     & 3.40                      & 3.44    \\ \cline{1-1}
age (only in   CLPsych 2015)    & $\sim$                   & $\sim$                   & $\sim$                    & $\sim$  & 45.19                    & 11.32                    & 23.40                     & 21.60   \\ \cline{1-1}
gender (only   in CLPsych 2015) & $\sim$                   & $\sim$                   & $\sim$                    & $\sim$  & 1.00                     & 0.00                     & 0.24                      & 0.00    \\ \hline
topic1  & 1.00 & 0.00 & 0.01 & 0.00 & 1.00 & 0.00 & 0.01 & 0.00 \\ \cline{1-1}
topic2  & 1.00 & 0.00 & 0.11 & 0.00 & 1.00 & 0.00 & 0.02 & 0.00 \\ \cline{1-1}
topic3  & 1.00 & 0.00 & 0.01 & 0.00 & 1.00 & 0.00 & 0.02 & 0.00 \\ \cline{1-1}
topic4  & 1.00 & 0.00 & 0.02 & 0.00 & 1.00 & 0.00 & 0.08 & 0.00 \\ \cline{1-1}
topic5  & 1.00 & 0.00 & 0.11 & 0.00 & 1.00 & 0.00 & 0.08 & 0.00 \\ \cline{1-1}
topic6  & 1.00 & 0.00 & 0.16 & 0.06 & 1.00 & 0.00 & 0.05 & 0.00 \\ \cline{1-1}
topic7  & 1.00 & 0.00 & 0.02 & 0.00 & 1.00 & 0.00 & 0.02 & 0.00 \\ \cline{1-1}
topic8  & 1.00 & 0.00 & 0.07 & 0.00 & 1.00 & 0.00 & 0.00 & 0.00 \\ \cline{1-1}
topic9  & 1.00 & 0.00 & 0.01 & 0.00 & 1.00 & 0.00 & 0.01 & 0.00 \\ \cline{1-1}
topic10 & 1.00 & 0.00 & 0.01 & 0.00 & 1.00 & 0.00 & 0.03 & 0.00 \\ \cline{1-1}
topic11 & 1.00 & 0.00 & 0.05 & 0.00 & 1.00 & 0.00 & 0.19 & 0.08 \\ \cline{1-1}
topic12 & 1.00 & 0.00 & 0.01 & 0.00 & 1.00 & 0.00 & 0.02 & 0.00 \\ \cline{1-1}
topic13 & 1.00 & 0.00 & 0.01 & 0.00 & 1.00 & 0.00 & 0.01 & 0.00 \\ \cline{1-1}
topic14 & 1.00 & 0.00 & 0.01 & 0.00 & 1.00 & 0.00 & 0.05 & 0.00 \\ \cline{1-1}
topic15 & 1.00 & 0.00 & 0.01 & 0.00 & 1.00 & 0.00 & 0.02 & 0.00 \\ \cline{1-1}
topic16 & 1.00 & 0.00 & 0.18 & 0.00 & 1.00 & 0.00 & 0.01 & 0.00 \\ \cline{1-1}
topic17 & 1.00 & 0.00 & 0.01 & 0.00 & 1.00 & 0.00 & 0.02 & 0.00 \\ \cline{1-1}
topic18 & 1.00 & 0.00 & 0.01 & 0.00 & 1.00 & 0.00 & 0.06 & 0.00 \\ \cline{1-1}
topic19 & 1.00 & 0.00 & 0.06 & 0.00 & 1.00 & 0.00 & 0.02 & 0.00 \\ \cline{1-1}
topic20 & 1.00 & 0.00 & 0.00 & 0.00 & 1.00 & 0.00 & 0.01 & 0.00 \\ \cline{1-1}
topic21 & 1.00 & 0.00 & 0.00 & 0.00 & 1.00 & 0.00 & 0.06 & 0.00 \\ \cline{1-1}
topic22 & 1.00 & 0.00 & 0.01 & 0.00 & 1.00 & 0.00 & 0.01 & 0.00 \\ \cline{1-1}
topic23 & 1.00 & 0.00 & 0.04 & 0.00 & 1.00 & 0.00 & 0.11 & 0.00 \\ \cline{1-1}
topic24 & 1.00 & 0.00 & 0.03 & 0.00 & 1.00 & 0.00 & 0.01 & 0.00 \\ \cline{1-1}
topic25 & 1.00 & 0.00 & 0.03 & 0.00 & 1.00 & 0.00 & 0.04 & 0.00 \\ \hline
\end{tabular}
\end{table}

\begin{table*}[htbp]
\centering
\tiny
\caption{Feature Statistics of individual classes for the two Twitter depression datasets.}
\label{Tab:feature_value2}
\begin{tabular}{|c|cccccccc|cccccccc|}
\hline
\multirow{3}{*}{\textbf{Features}}       & \multicolumn{8}{c|}{\textbf{TTDD}}                                                                                                                                                                                 & \multicolumn{8}{c|}{\textbf{CLPsych 2015}}                                                                                                                                                                          \\ \cline{2-17} 
                                & \multicolumn{4}{c|}{Depressed}                                                                                & \multicolumn{4}{c|}{Control}                                                              & \multicolumn{4}{c|}{Depressed}                                                                                 & \multicolumn{4}{c|}{Control}                                                              \\ \cline{2-17} 
                                & \multicolumn{1}{c|}{max} & \multicolumn{1}{c|}{min} & \multicolumn{1}{c|}{mean} & \multicolumn{1}{c|}{median} & \multicolumn{1}{c|}{max} & \multicolumn{1}{c|}{min} & \multicolumn{1}{c|}{mean} & median  & \multicolumn{1}{c|}{max} & \multicolumn{1}{c|}{min} & \multicolumn{1}{c|}{mean} & \multicolumn{1}{c|}{median}  & \multicolumn{1}{c|}{max} & \multicolumn{1}{c|}{min} & \multicolumn{1}{c|}{mean} & median  \\ \hline
total   favourites              & 84257.00                 & 0.00                     & 2823.50                   & \multicolumn{1}{c|}{698.00} & 356830.00                & 0.00                     & 8709.07                   & 2341.00 & 226236.00                & 0.00                     & 3849.46                   & \multicolumn{1}{c|}{925.62}  & 50162.00                 & 0.00                     & 1190.48                   & 182.50  \\ \cline{1-1}
friends                         & 61572.00                 & 0.00                     & 745.17                    & \multicolumn{1}{c|}{307.00} & 11977.00                 & 0.00                     & 643.16                    & 340.00  & 72296.00                 & 1.00                     & 1205.13                   & \multicolumn{1}{c|}{369.00}  & 21055.00                 & 0.00                     & 654.05                    & 302.00  \\ \cline{1-1}
followers                       & 102069.00                & 0.00                     & 1062.63                   & \multicolumn{1}{c|}{257.00} & 14829.00                 & 0.00                     & 1027.27                   & 434.00  & 154744.00                & 0.00                     & 2167.23                   & \multicolumn{1}{c|}{411.10}  & 234328.00                & 0.00                     & 1513.29                   & 292.50  \\ \cline{1-1}
listed count                    & 1575.00                  & 0.00                     & 16.01                     & \multicolumn{1}{c|}{3.00}   & 1722.00                  & 0.00                     & 18.98                     & 4.00    & 1359.00                  & 0.00                     & 19.44                     & \multicolumn{1}{c|}{2.00}    & 2156.00                  & 0.00                     & 19.35                     & 1.00    \\ \cline{1-1}
favourites   count              & 1449.55                  & 0.00                     & 1.09                      & \multicolumn{1}{c|}{0.19}   & 67.33                    & 0.00                     & 0.81                      & 0.29    & 37.59                    & 0.00                     & 0.49                      & \multicolumn{1}{c|}{0.16}    & 127.77                   & 0.00                     & 0.32                      & 0.07    \\ \cline{1-1}
retweet count                   & 126999.12                & 0.00                     & 925.97                    & \multicolumn{1}{c|}{64.39}  & 169788.00                & 0.00                     & 2332.88                   & 649.44  & 24261.07                 & 0.00                     & 892.42                    & \multicolumn{1}{c|}{139.09}  & 40743.26                 & 0.00                     & 470.37                    & 63.14   \\ \cline{1-1}
mention count                   & 3.00                     & 0.00                     & 0.60                      & \multicolumn{1}{c|}{0.56}   & 5.52                     & 0.00                     & 0.74                      & 0.72    & 2.60                     & 0.00                     & 0.67                      & \multicolumn{1}{c|}{0.62}    & 2.68                     & 0.00                     & 0.70                      & 0.65    \\ \cline{1-1}
posting number                  & 2560.00                  & 6.00                     & 187.99                    & \multicolumn{1}{c|}{104.00} & 2972.00                  & 6.00                     & 601.11                    & 274.00  & 3000.00                  & 15.00                    & 2202.70                   & \multicolumn{1}{c|}{2711.50} & 3000.00                  & 23.00                    & 2051.03                   & 2641.00 \\ \cline{1-1}
time   distribution             & 1.00                     & 0.00                     & 0.35                      & \multicolumn{1}{c|}{0.35}   & 0.97                     & 0.00                     & 0.29                      & 0.30    & 1.00                     & 0.00                     & 0.36                      & \multicolumn{1}{c|}{0.39}    & 0.81                     & 0.00                     & 0.34                      & 0.37    \\ \cline{1-1}
negative words                  & 3.33                     & 0.07                     & 0.81                      & \multicolumn{1}{c|}{0.76}   & 3.68                     & 0.00                     & 0.62                      & 0.61    & 2.01                     & 0.07                     & 0.71                      & \multicolumn{1}{c|}{0.68}    & 1.77                     & 0.00                     & 0.63                      & 0.61    \\ \cline{1-1}
positive words                  & 4.33                     & 0.29                     & 1.47                      & \multicolumn{1}{c|}{1.41}   & 5.50                     & 0.00                     & 1.22                      & 1.20    & 2.57                     & 0.14                     & 1.30                      & \multicolumn{1}{c|}{1.26}    & 3.05                     & 0.00                     & 1.21                      & 1.19    \\ \cline{1-1}
emoji number                    & 8.59                     & 0.00                     & 0.23                      & \multicolumn{1}{c|}{0.06}   & 9.31                     & 0.00                     & 0.40                      & 0.25    & 0.00                     & 0.00                     & 0.00                      & \multicolumn{1}{c|}{0.00}    & 0.00                     & 0.00                     & 0.00                      & 0.00    \\ \cline{1-1}
emoticons                       & 1.39                     & 0.00                     & 0.27                      & \multicolumn{1}{c|}{0.24}   & 3.00                     & 0.00                     & 0.25                      & 0.22    & 4.00                     & 2.28                     & 3.43                      & \multicolumn{1}{c|}{3.47}    & 4.00                     & 1.11                     & 3.37                      & 3.42    \\ \cline{1-1}
age (only in   CLPsych 2015)    & $\sim$                   & $\sim$                   & $\sim$                    & \multicolumn{1}{c|}{$\sim$} & $\sim$                   & $\sim$                   & $\sim$                    & $\sim$  & 40.49                    & 13.28                    & 21.66                     & \multicolumn{1}{c|}{20.07}   & 45.19                    & 11.32                    & 24.34                     & 22.80   \\ \cline{1-1}
gender (only   in CLPsych 2015) & $\sim$                   & $\sim$                   & $\sim$                    & \multicolumn{1}{c|}{$\sim$} & $\sim$                   & $\sim$                   & $\sim$                    & $\sim$  & 1.00                     & 0.00                     & 0.20                      & \multicolumn{1}{c|}{0.00}    & 1.00                     & 0.00                     & 0.25                      & 0.00    \\ \hline
topic1  & 0.91 & 0.00 & 0.00 &  \multicolumn{1}{c|}{0.00} & 1.00 & 0.00 & 0.01 & 0.00 & 0.45 & 0.00 & 0.01 &  \multicolumn{1}{c|}{0.00} & 1.00 & 0.00 & 0.01 & 0.00 \\ \cline{1-1}
topic2  & 1.00 & 0.00 & 0.24 &  \multicolumn{1}{c|}{0.15} & 1.00 & 0.00 & 0.04 & 0.00 & 1.00 & 0.00 & 0.02 & \multicolumn{1}{c|}{ 0.00} & 1.00 & 0.00 & 0.01 & 0.00 \\ \cline{1-1}
topic3  & 0.99 & 0.00 & 0.01 &  \multicolumn{1}{c|}{0.00} & 1.00 & 0.00 & 0.01 & 0.00 & 0.73 & 0.00 & 0.01 &  \multicolumn{1}{c|}{0.00} & 1.00 & 0.00 & 0.03 & 0.00 \\ \cline{1-1}
topic4  & 0.84 & 0.00 & 0.01 & \multicolumn{1}{c|}{ 0.00} & 1.00 & 0.00 & 0.02 & 0.00 & 0.86 & 0.00 & 0.05 &  \multicolumn{1}{c|}{0.00} & 1.00 & 0.00 & 0.11 & 0.00 \\ \cline{1-1}
topic5  & 0.94 & 0.00 & 0.10 &  \multicolumn{1}{c|}{0.00} & 1.00 & 0.00 & 0.12 & 0.00 & 1.00 & 0.00 & 0.11 &  \multicolumn{1}{c|}{0.00} & 0.96 & 0.00 & 0.05 & 0.00 \\ \cline{1-1}
topic6  & 0.99 & 0.00 & 0.20 &  \multicolumn{1}{c|}{0.15} & 1.00 & 0.00 & 0.13 & 0.03 & 0.91 & 0.00 & 0.04 &  \multicolumn{1}{c|}{0.00} & 1.00 & 0.00 & 0.06 & 0.00 \\ \cline{1-1}
topic7  & 1.00 & 0.00 & 0.01 &  \multicolumn{1}{c|}{0.00} & 1.00 & 0.00 & 0.02 & 0.00 & 0.99 & 0.00 & 0.03 &  \multicolumn{1}{c|}{0.00} & 1.00 & 0.00 & 0.02 & 0.00 \\ \cline{1-1}
topic8  & 0.99 & 0.00 & 0.11 &  \multicolumn{1}{c|}{0.00} & 1.00 & 0.00 & 0.06 & 0.00 & 0.74 & 0.00 & 0.01 &  \multicolumn{1}{c|}{0.00} & 1.00 & 0.00 & 0.01 & 0.00 \\ \cline{1-1}
topic9  & 0.96 & 0.00 & 0.00 &  \multicolumn{1}{c|}{0.00} & 1.00 & 0.00 & 0.02 & 0.00 & 0.57 & 0.00 & 0.00 &  \multicolumn{1}{c|}{0.00} & 1.00 & 0.00 & 0.01 & 0.00 \\ \cline{1-1}
topic10 & 0.98 & 0.00 & 0.02 &  \multicolumn{1}{c|}{0.00} & 1.00 & 0.00 & 0.01 & 0.00 & 0.79 & 0.00 & 0.01 &  \multicolumn{1}{c|}{0.00} & 1.00 & 0.00 & 0.04 & 0.00 \\ \cline{1-1}
topic11 & 0.89 & 0.00 & 0.03 &  \multicolumn{1}{c|}{0.00} & 1.00 & 0.00 & 0.06 & 0.00 & 1.00 & 0.00 & 0.27 &  \multicolumn{1}{c|}{0.20} & 0.98 & 0.00 & 0.15 & 0.02 \\ \cline{1-1}
topic12 & 0.52 & 0.00 & 0.01 &  \multicolumn{1}{c|}{0.00} & 1.00 & 0.00 & 0.01 & 0.00 & 1.00 & 0.00 & 0.06 &  \multicolumn{1}{c|}{0.00} & 0.97 & 0.00 & 0.01 & 0.00 \\ \cline{1-1}
topic13 & 0.65 & 0.00 & 0.00 &  \multicolumn{1}{c|}{0.00} & 1.00 & 0.00 & 0.01 & 0.00 & 0.95 & 0.00 & 0.01 &  \multicolumn{1}{c|}{0.00} & 1.00 & 0.00 & 0.02 & 0.00 \\ \cline{1-1}
topic14 & 0.61 & 0.00 & 0.01 &  \multicolumn{1}{c|}{0.00} & 1.00 & 0.00 & 0.01 & 0.00 & 0.95 & 0.00 & 0.03 &  \multicolumn{1}{c|}{0.00} & 1.00 & 0.00 & 0.04 & 0.00 \\ \cline{1-1}
topic15 & 0.97 & 0.00 & 0.02 &  \multicolumn{1}{c|}{0.00} & 1.00 & 0.00 & 0.01 & 0.00 & 1.00 & 0.00 & 0.02 & \multicolumn{1}{c|}{ 0.00} & 1.00 & 0.00 & 0.03 & 0.00 \\ \cline{1-1}
topic16 & 0.99 & 0.00 & 0.08 &  \multicolumn{1}{c|}{0.00} & 1.00 & 0.00 & 0.23 & 0.03 & 0.75 & 0.00 & 0.00 &  \multicolumn{1}{c|}{0.00} & 1.00 & 0.00 & 0.01 & 0.00 \\ \cline{1-1}
topic17 & 0.74 & 0.00 & 0.01 &  \multicolumn{1}{c|}{0.00} & 1.00 & 0.00 & 0.02 & 0.00 & 0.96 & 0.00 & 0.01 &  \multicolumn{1}{c|}{0.00} & 1.00 & 0.00 & 0.01 & 0.00 \\ \cline{1-1}
topic18 & 0.86 & 0.00 & 0.01 &  \multicolumn{1}{c|}{0.00} & 1.00 & 0.00 & 0.01 & 0.00 & 0.97 & 0.00 & 0.05 &  \multicolumn{1}{c|}{0.00} & 1.00 & 0.00 & 0.08 & 0.00 \\ \cline{1-1}
topic19 & 0.99 & 0.00 & 0.07 &  \multicolumn{1}{c|}{0.00} & 1.00 & 0.00 & 0.06 & 0.00 & 0.59 & 0.00 & 0.01 &  \multicolumn{1}{c|}{0.00} & 1.00 & 0.00 & 0.03 & 0.00 \\ \cline{1-1}
topic20 & 0.24 & 0.00 & 0.00 &  \multicolumn{1}{c|}{0.00} & 1.00 & 0.00 & 0.00 & 0.00 & 1.00 & 0.00 & 0.01 &  \multicolumn{1}{c|}{0.00} & 1.00 & 0.00 & 0.01 & 0.00 \\ \cline{1-1}
topic21 & 0.49 & 0.00 & 0.00 &  \multicolumn{1}{c|}{0.00} & 1.00 & 0.00 & 0.01 & 0.00 & 1.00 & 0.00 & 0.10 &  \multicolumn{1}{c|}{0.00} & 1.00 & 0.00 & 0.05 & 0.00 \\ \cline{1-1}
topic22 & 0.28 & 0.00 & 0.00 &  \multicolumn{1}{c|}{0.00} & 1.00 & 0.00 & 0.01 & 0.00 & 0.46 & 0.00 & 0.00 &  \multicolumn{1}{c|}{0.00} & 1.00 & 0.00 & 0.02 & 0.00 \\ \cline{1-1}
topic23 & 0.74 & 0.00 & 0.02 &  \multicolumn{1}{c|}{0.00} & 1.00 & 0.00 & 0.05 & 0.00 & 1.00 & 0.00 & 0.10 &  \multicolumn{1}{c|}{0.00} & 1.00 & 0.00 & 0.16 & 0.00 \\ \cline{1-1}
topic24 & 0.64 & 0.00 & 0.01 &  \multicolumn{1}{c|}{0.00} & 1.00 & 0.00 & 0.04 & 0.00 & 0.84 & 0.00 & 0.00 &  \multicolumn{1}{c|}{0.00} & 1.00 & 0.00 & 0.01 & 0.00 \\ \cline{1-1}
topic25 & 0.73 & 0.00 & 0.00 &  \multicolumn{1}{c|}{0.00} & 1.00 & 0.00 & 0.04 & 0.00 & 1.00 & 0.00 & 0.02 & \multicolumn{1}{c|}{ 0.00} & 0.88 & 0.00 & 0.02 & 0.00 \\ \hline
\end{tabular}
\end{table*}

\begin{landscape}
\begin{table*}
\footnotesize
\centering
\makeatletter\def\@captype{table}\makeatother\caption{{Summary of methods for Social networks users depression classification.}}
\begin{tabular}{lllll}
\hline
\rowcolor[rgb]{0.753,0.753,0.753} Authors & Extracted Features                                                                                                                                      & Critical Features                                                                                & Employed Classifier                                                                                                 & Limitations                                                                                                                                                                                                            \\
\hline
Park et al.            \cite{park2013perception}                   & \begin{tabular}[c]{@{}l@{}}Sentence polarity, user~\\profiles, posting time \\distribution......\\(totally 7 features)\end{tabular}                        & Sentence polarity                                                                                & LR                                                                                                                  & \begin{tabular}[c]{@{}l@{}}Extracted features are \\not comprehensive. \end{tabular}                                                                                                                                   \\
\hline
Nadeem et al.      \cite{nadeem2016identifying}                       & \begin{tabular}[c]{@{}l@{}}HDP topics, BoW vectors,\\posting time, user profiles......\\(totally about 20 features)\end{tabular}                             & HDP topics                                                                                       & \begin{tabular}[c]{@{}l@{}}Four binary classifiers,\\ i.e. SVM, LR,\\Naive Bayes \\ and Decision tree \end{tabular} & \begin{tabular}[c]{@{}l@{}}These classifiers have a \\poor fitness ability to \\handle complex dataset.~\end{tabular}                                                                                                    \\
\hline
Choudhury et al.  \cite{de2013social}                     & \begin{tabular}[c]{@{}l@{}}user profiles,~\\negative word count, \\BoW, LIWC,~followers \\and friends number...... \\(totally about 35 features)\end{tabular} & \begin{tabular}[c]{@{}l@{}}LIWC \\sentence polarity,\\cognitive score, \\BoW vector\end{tabular} & SVM with RBF kernal                                                                                                 & \begin{tabular}[c]{@{}l@{}}It is easy to be disturbed \\by noise information.\end{tabular}                                                                                                                              \\
\hline
Shuai et al.          \cite{shuai2018comprehensive}                    & \begin{tabular}[c]{@{}l@{}}LDA topics, \\user profile information......\\(totally about 25 features)\end{tabular}                                             & LDA topics                                                                                       & \begin{tabular}[c]{@{}l@{}}Ensemble model \\combine\\ SVM and LR \end{tabular}                                      & \begin{tabular}[c]{@{}l@{}}This model is not robust \\and stable~because \\the employed classifier \\does not have strong fitting \\ability in complex datasets.\end{tabular}                                          \\
\hline
Jamil et al.         \cite{jamil2017monitoring}                     & \begin{tabular}[c]{@{}l@{}}BoW, posting count, posting time......\\(totally about 10 features)\end{tabular}                                                   & BoW                                                                                              & PCA+SVM                                                                                                             & \begin{tabular}[c]{@{}l@{}}Lack of comprehensively \\feature mining from \\posting texts.\end{tabular}                                                                                                                   \\
\hline
Song et al.  \cite{song2015multiple}                          & \begin{tabular}[c]{@{}l@{}}LIWC, LSA topics, users profiles,\\favourite posting count......\\(totally about 30 features)\end{tabular}                         & \begin{tabular}[c]{@{}l@{}}LIWC features, \\LSA topics\end{tabular}                              & \begin{tabular}[c]{@{}l@{}}Multi-task learning\\+SVM\end{tabular}                                                   & \begin{tabular}[c]{@{}l@{}}It has not been evaluated \\on publicly accessible online \\depression detection \\datasets.\end{tabular}                                                                                           \\
\hline
Shen et al.       \cite{shen2017depression}                        & \begin{tabular}[c]{@{}l@{}}Six groups' feature like~\\user engagement and \\domain-specific keyword......\\(totally 35 features)\end{tabular}                 & \begin{tabular}[c]{@{}l@{}}LDA topics, \\domain-specific \\keywords\end{tabular}                 & \begin{tabular}[c]{@{}l@{}}Multimodal dictionary \\ learning + LR \end{tabular}                                     & \begin{tabular}[c]{@{}l@{}}It shares the weaknesses\\ with MOD being efficient\\ only for signals with relatively\\ low dimensionality and having \\ the possibility of being stuck \\ at local minima. \end{tabular}  \\
\hline
Cacheda et al.  \cite{cacheda2019early}                          & \begin{tabular}[c]{@{}l@{}}LSA topics, posting time, \\negative word count......\\(totally about 15 feature)\end{tabular}                                     & LSA topics                                                                                       & Random Forest                                                                                                       & \begin{tabular}[c]{@{}l@{}}It is easy to be disturbed~\\by noise information and \\lack of~making contributions \\on universal algorithm.\end{tabular}                                                                  \\
\hline
Shen et al.  \cite{shen2018cross}                          & \begin{tabular}[c]{@{}l@{}}Emotional word count,\\ posting time, \\Tweet count......\\(totally about 60 features)\end{tabular}                                     & $\setminus$ & Multi-layer Perceptron
 & \begin{tabular}[c]{@{}l@{}}This method aims to improve the\\ recognition performance in  Weibo\\ domain utilising the rich data  of \\ Twitter but the classification  accuracy in the \\ Twitter dataset is not good.
 \end{tabular}                                                                  \\
\hline
Ray et al.  \cite{ray2019multi}                          & \begin{tabular}[c]{@{}l@{}}Transcript context features, \\audio context features\\Visual context features......\\(3 feature categories with 400000 dimensions)\end{tabular}                                     & $\setminus$ & Multi-level attention network & \begin{tabular}[c]{@{}l@{}}Need extra information like\\the facial emotion and\\audio of the depressed people. \end{tabular}                                                                  \\
\hline
Gamaarachchige et al.  \cite{gamaarachchige2019multi}                          & \begin{tabular}[c]{@{}l@{}}Text tokens, TF-IDF scores, \\emotion label......\\(6 feature categories with 46200 dimensions)\end{tabular}                                     & Text tokens                                                                                       & \begin{tabular}[c]{@{}l@{}} Multi-task, multi-channel,\\ multi-input model\end{tabular}    
& \begin{tabular}[c]{@{}l@{}}This method lacks clear interpretations \\ to the model predictions as \\ of which specific factor \\ influences the predicted \\ depression risk.\end{tabular}                                                                  \\
\hline
Orabi et al.  \cite{orabi2018deep}                          & \begin{tabular}[c]{@{}l@{}}Word embeddings like \\Skip-Gram, CBOW......\\(300 dimensional word embeddings)\end{tabular}                                     & Word embeddings & CNN-LSTM network & \begin{tabular}[c]{@{}l@{}}Only texture features have \\ been extracted. Users' online \\ behaviours cannot be \\ well-described.\end{tabular}                                                                  \\
\hline
Ours                                      & \begin{tabular}[c]{@{}l@{}}Three features categories \\including user profile features,~\\user social interaction features\\and linguistic features......\\ (totally 14 features)\end{tabular}                            & \begin{tabular}[c]{@{}l@{}}LDA topics \end{tabular}       & \begin{tabular}[c]{@{}l@{}}CBPT\end{tabular}                                          & \begin{tabular}[c]{@{}l@{}}Longer training time \\compared with others traditional\\ machine learning based classifiers.\end{tabular}                                                                                                                                   \\
\hline
\end{tabular}

\label{Tab:literature_compare}
\end{table*}
\end{landscape}

\newpage
\section{Supplementary B: Settings of the Datasets and the Boosting Methods}
\textbf{UCI Dataset}: (1) LSVT Dataset. This dataset collected the clinical information from the speech signals and contained 128 samples, 309 features and 2 classes \cite{tsanas2014objective}. (2) Statlog Dataset. This dataset harvested the information of the statlog and included 6435 samples, 37 features, and 2 classes \cite{fernandes2015proactive}. (3) Glass Dataset. This dataset collected information of glasses left in the crime scene and contains 214 samples, 10 features, and 6 classes \cite{bredensteiner1999multicategory}. The datasets' information is summarised in Table \ref{Tab:dataset_sum}.

Table \ref{Tab:lightGBM}-10 shows the parameter settings for the boosting algorithms.

\begin{table*}[htbp]
\centering
\caption{Experiment Datasets.}
\label{Tab:dataset_sum}
\begin{tabular}{|l|c|c|c|c|c|}
\hline
Dataset Scale\textbackslash{}Name & \multicolumn{1}{l|}{TTDD} & \multicolumn{1}{l|}{CLPsych 2015} & \multicolumn{1}{l|}{LSVT} & \multicolumn{1}{l|}{Statlog} & \multicolumn{1}{l|}{Glass} \\ \hline
Total Samples                     & 7873                      & 1746                              & 128                       & 6435                         & 214                        \\ \hline
Feature Dimension                 & 38                        & 40                                & 309                       & 37                           & 10                         \\ \hline
Class number                      & 2                         & 3                                 & 2                         & 6                            & 6                          \\ \hline
\end{tabular}
\end{table*}

\begin{table*}[htbp]
\centering
        \begin{minipage}[h]{0.3\linewidth}
         \makeatletter\def\@captype{table}\makeatother\caption{Parameter Setting for LightGBM}
\scalebox{1}[1]{
\begin{tabular}{|l|c|}
\hline
\textit{num leaves}    & 64, 128, 256 \\ \hline
\textit{max bins}      & 63,255       \\ \hline
\textit{max depth}     & 5,10,15      \\ \hline
\textit{learning rate} & 0.1, 0.5,1   \\ \hline
\textit{reg lambda}    & 0.1,1        \\ \hline
\textit{tree number}   & 500          \\ \hline
\end{tabular}
}

            \label{Tab:lightGBM}
        \end{minipage}
        \hfill
        \begin{minipage}[h]{0.3\linewidth}
         \makeatletter\def\@captype{table}\makeatother\caption{Parameter Setting for XGboost}
 \scalebox{1}[1]{
\begin{tabular}{|l|c|}
\hline
\textit{num leaves}    & 64, 128, 256 \\ \hline
\textit{max bins}      & 63,255       \\ \hline
\textit{max depth}     & 5,10,15      \\ \hline
\textit{eta} & 0.1, 0.5,1   \\ \hline
\textit{lambda}    & 0.1,1        \\ \hline
\textit{tree number}   & 500          \\ \hline
\end{tabular}
}

             \label{Tab:LSVT_set}
        \end{minipage}
        \hfill
               \begin{minipage}[h]{0.3\linewidth}
        \makeatletter\def\@captype{table}\makeatother\caption{Parameter Setting for KiGB.}
\scalebox{1}[1]{
\begin{tabular}{|l|c|}
\hline
\textit{num leaves}    & 64, 128, 256 \\ \hline
\textit{loss}      & 'deviance'      \\ \hline
\textit{max depth}     & 5,10,15      \\ \hline
\textit{learning rate} & 0.1, 0.5,1   \\ \hline
\textit{min sample split}    & 2,5        \\ \hline
\textit{tree number}   & 500          \\ \hline
\end{tabular}
}

            \label{Tab:KiGB}
        \end{minipage}

              \begin{minipage}[h]{0.3\linewidth}
         \makeatletter\def\@captype{table}\makeatother\caption{Parameter Setting for Discrete Adaboost}
\scalebox{1}[1]{
\begin{tabular}{|l|c|}
\hline
\textit{num leaves}    & - \\ \hline
\textit{algorithm}      & 'SAMME'       \\ \hline
\textit{max depth}     & -      \\ \hline
\textit{learning rate} & 0.1, 0.5,1   \\ \hline
\textit{tree number}   & 500          \\ \hline
\end{tabular}
}

            \
        \end{minipage}
        \hfill
                \begin{minipage}[h]{0.3\linewidth}
         \makeatletter\def\@captype{table}\makeatother\caption{Parameter Setting for Real Adaboost}
\scalebox{1}[1]{
\begin{tabular}{|l|c|}
\hline
\textit{num leaves}    & - \\ \hline
\textit{algorithm}      & 'SAMME.R'      \\ \hline
\textit{max depth}     & -     \\ \hline
\textit{learning rate} & 0.1, 0.5,1   \\ \hline
\textit{tree number}   & 500          \\ \hline
\end{tabular}
}

        \end{minipage}
        \hfill
                \begin{minipage}[h]{0.3\linewidth}
         \makeatletter\def\@captype{table}\makeatother\caption{Parameter Setting for LogitBoost}
\scalebox{1}[1]{
\begin{tabular}{|l|c|}
\hline
\textit{num leaves}    & 64, 128, 256 \\ \hline
\textit{weight trim quantile}      & 0.05,0.1,0.5       \\ \hline
\textit{max depth}     & 5,10,15      \\ \hline
\textit{learning rate} & 0.1, 0.5,1   \\ \hline
\textit{max response}    & 2,4,8       \\ \hline
\textit{tree number}   & 500          \\ \hline
\end{tabular}
}

        \end{minipage}

                \begin{minipage}[h]{0.3\linewidth}
         \makeatletter\def\@captype{table}\makeatother\caption{Parameter Setting for CBPT}
\scalebox{1}[1]{
\begin{tabular}{|l|c|}
\hline
\textit{penalty terms coefficient}      & 0.5      \\ \hline
\textit{data scaling low limit}      & 1      \\ \hline
\textit{learning rate} & 0.1, 0.5,1   \\ \hline
\textit{tree number}   & 500          \\ \hline
\textit{resampling times}   & 5          \\ \hline
\end{tabular}
}

        \end{minipage}

\end{table*}

\newpage
\section{Supplementary C: TreeSHAP Plots}
Tables \ref{Tab:lda_ttdd} and \ref{Tab:lda_clp} show the LDA topic model results with top 10 words. In order to explain the meaning of the topic features, we referring to the literatures \cite{miao2016neural,moody2016mixing} and deduce the corresponding topics based on the top words. The deduced topics are shown in the second column of tables. For example, topic1 of TTDD is inferred as 'Youtube Video' as it includes words such as 'earning', 'youtube', 'link'. Some confusing topics like topic20 with messy words are labelled as '$\sim$'.

Fig. \ref{fig:feature_importances} shows the feature importance of the two Twitter datasets. 

\begin{figure*}[htbp]
     \centering
     \begin{subfigure}[b]{0.32\textwidth}
         \centering
         \includegraphics[width=1\textwidth]{figures/ttdd_dependence_emojis.png}

     \end{subfigure}
     \hfill
     \begin{subfigure}[b]{0.32\textwidth}
         \centering
         \includegraphics[width=1\textwidth]{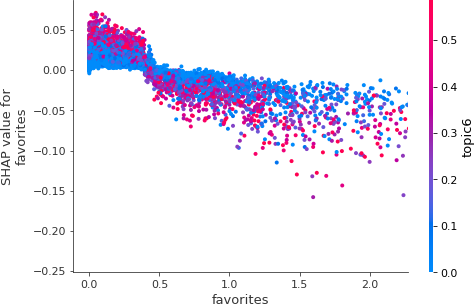}

     \end{subfigure}
     \hfill
     \begin{subfigure}[b]{0.32\textwidth}
         \centering
         \includegraphics[width=1\textwidth]{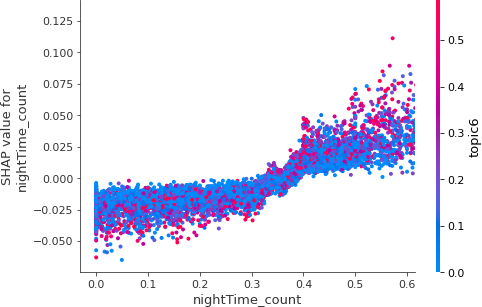}

     \end{subfigure}
     \hfill
     \begin{subfigure}[b]{0.32\textwidth}
         \centering
         \includegraphics[width=1\textwidth]{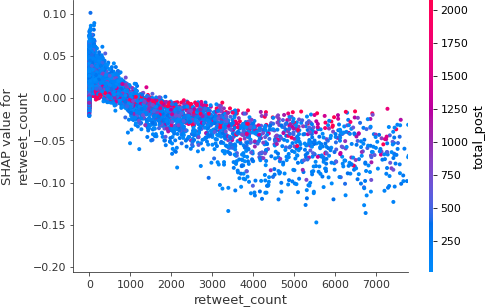}

     \end{subfigure}
     \hfill
          \begin{subfigure}[b]{0.32\textwidth}
         \centering
         \includegraphics[width=1\textwidth]{figures/ttdd_dependence_topic2.png}

     \end{subfigure}
     \hfill
          \begin{subfigure}[b]{0.32\textwidth}
         \centering
         \includegraphics[width=1\textwidth]{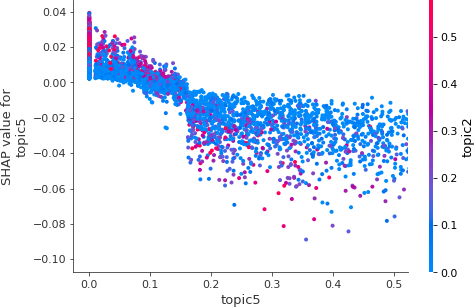}

     \end{subfigure}
     \hfill
               \begin{subfigure}[b]{0.32\textwidth}
         \centering
         \includegraphics[width=1\textwidth]{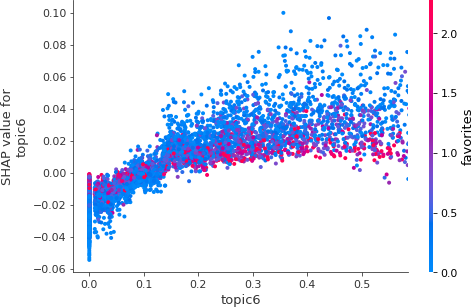}

     \end{subfigure}
     \hfill
               \begin{subfigure}[b]{0.32\textwidth}
         \centering
         \includegraphics[width=1\textwidth]{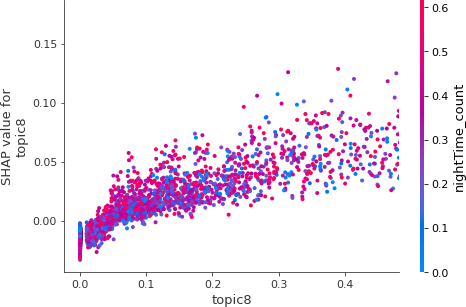}

     \end{subfigure}
     \hfill
               \begin{subfigure}[b]{0.32\textwidth}
         \centering
         \includegraphics[width=1\textwidth]{figures/ttdd_dependence_total_post.png}

     \end{subfigure}
     \hfill

        \caption{Top 9 significant feature dependency for TTDD dataset.}
        \label{fig:ttdd_dp}
\end{figure*}

\begin{figure*}[htbp]
     \centering
     \begin{subfigure}[b]{0.32\textwidth}
         \centering
         \includegraphics[width=1\textwidth]{figures/clp_dependence_age.png}

     \end{subfigure}
     \hfill
     \begin{subfigure}[b]{0.32\textwidth}
         \centering
         \includegraphics[width=1\textwidth]{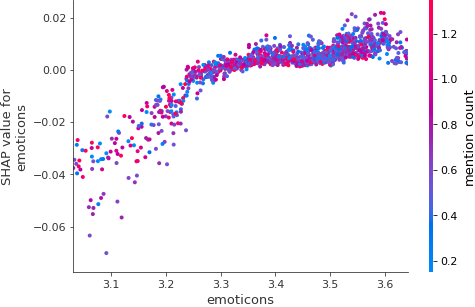}

     \end{subfigure}
     \hfill
     \begin{subfigure}[b]{0.32\textwidth}
         \centering
         \includegraphics[width=1\textwidth]{figures/clp_dependence_favorites.png}

     \end{subfigure}
     \hfill
     \begin{subfigure}[b]{0.32\textwidth}
         \centering
         \includegraphics[width=1\textwidth]{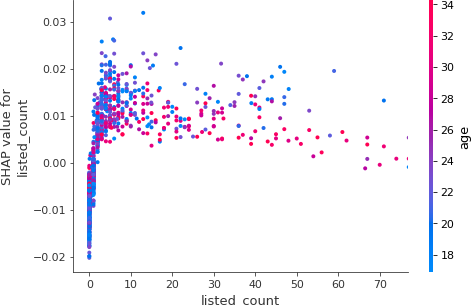}

     \end{subfigure}
     \hfill
          \begin{subfigure}[b]{0.32\textwidth}
         \centering
         \includegraphics[width=1\textwidth]{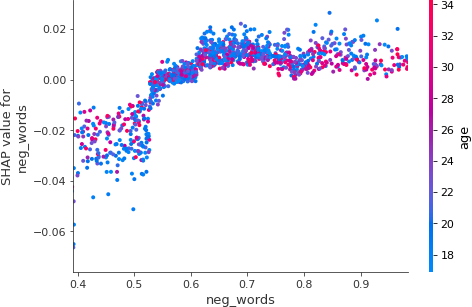}

     \end{subfigure}
     \hfill
          \begin{subfigure}[b]{0.32\textwidth}
         \centering
         \includegraphics[width=1\textwidth]{figures/clp_dependence_topic4.png}

     \end{subfigure}
     \hfill
               \begin{subfigure}[b]{0.32\textwidth}
         \centering
         \includegraphics[width=1\textwidth]{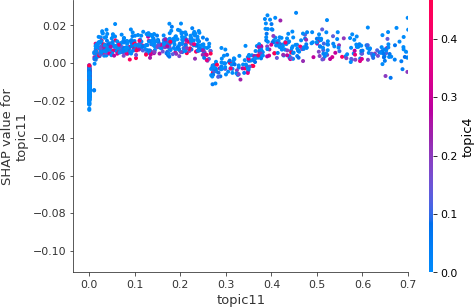}

     \end{subfigure}
     \hfill
               \begin{subfigure}[b]{0.32\textwidth}
         \centering
         \includegraphics[width=1\textwidth]{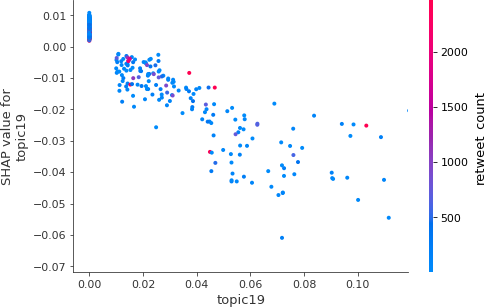}

     \end{subfigure}
     \hfill
               \begin{subfigure}[b]{0.32\textwidth}
         \centering
         \includegraphics[width=1\textwidth]{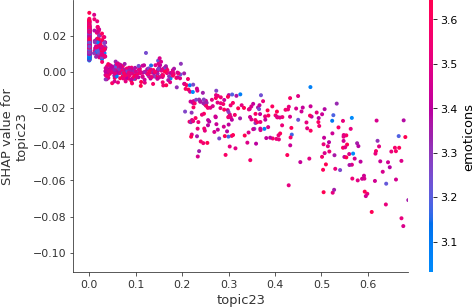}

     \end{subfigure}
     \hfill

        \caption{Top 9 significant feature dependency for CLPsych 2015 CvD subset.}
        \label{fig:clp_dp}
\end{figure*}

Fig. \ref{fig:ttdd_dp}. and \ref{fig:clp_dp} shows the dependency plots of top 9 significant features.

\begin{figure*}[htbp]
     \centering
     \begin{subfigure}[b]{0.48\textwidth}
         \centering
         \includegraphics[width=1\textwidth]{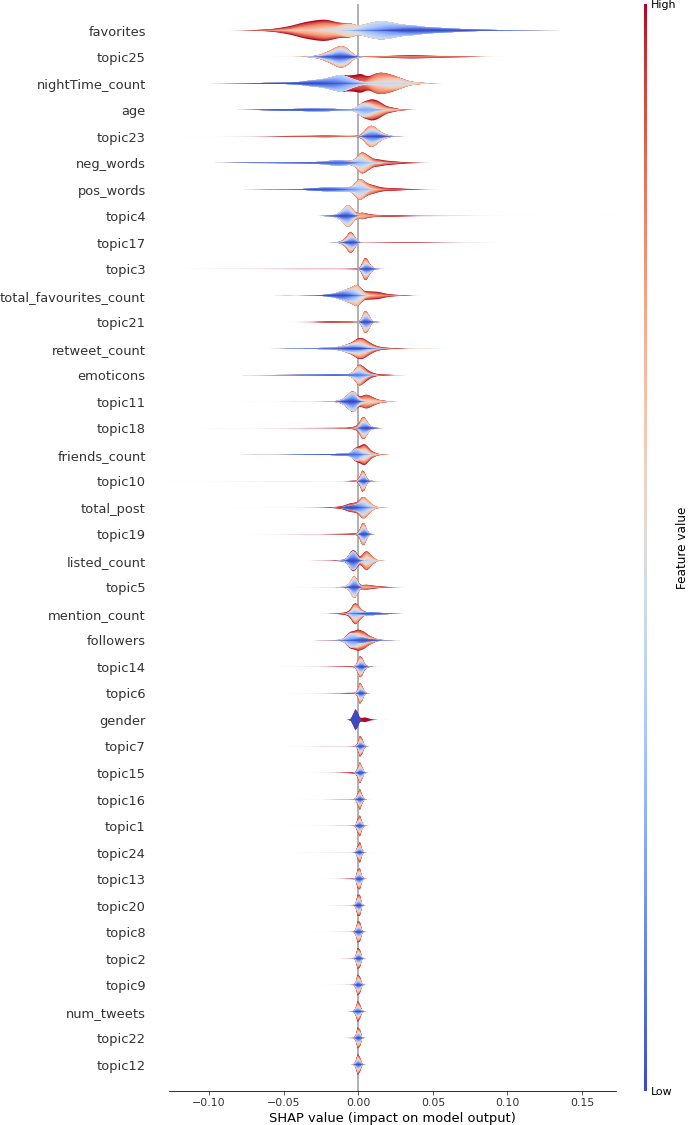}
         \caption{CLPsych 2015 CvP Local: Impact of each feature on model prediction.}
		
     \end{subfigure}
     \hfill
     \begin{subfigure}[b]{0.48\textwidth}
         \centering
         \includegraphics[width=1\textwidth]{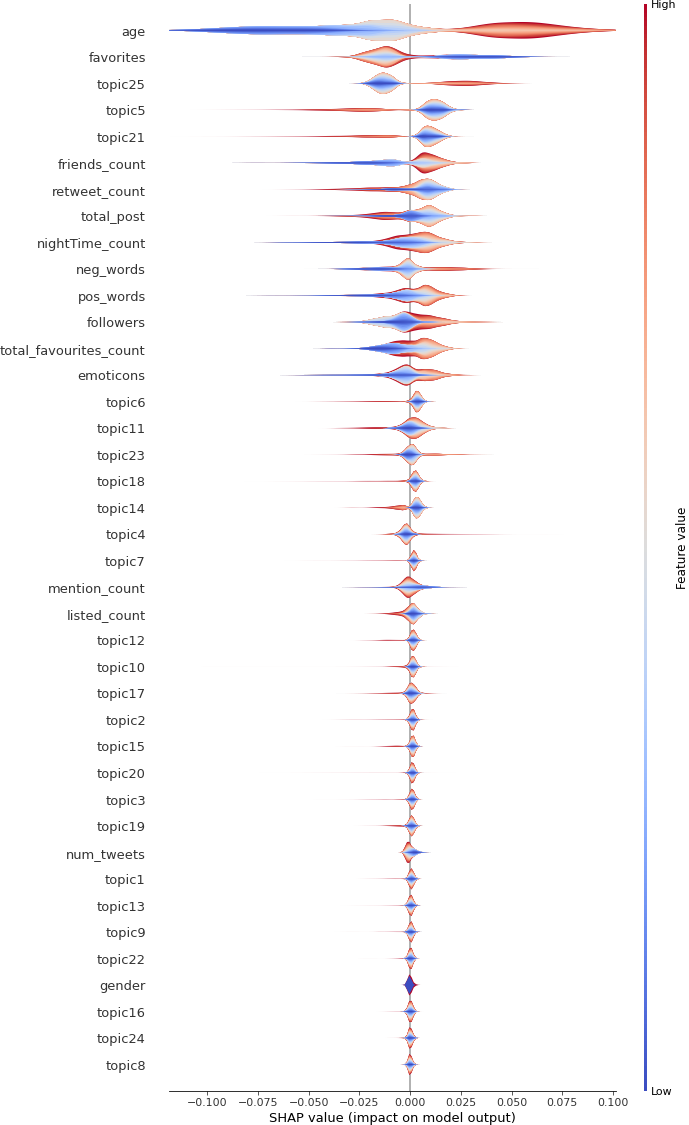}
         \caption{CLPsych 2015 PvD subset: Impact of each feature on model prediction.}
     \end{subfigure}
     \hfill

        \caption{Feature Importance.}
        \label{fig:feature_importances2}
\end{figure*}

\begin{figure*}
     \centering
     \begin{subfigure}[b]{1\textwidth}
         \centering
         \includegraphics[width=1\textwidth]{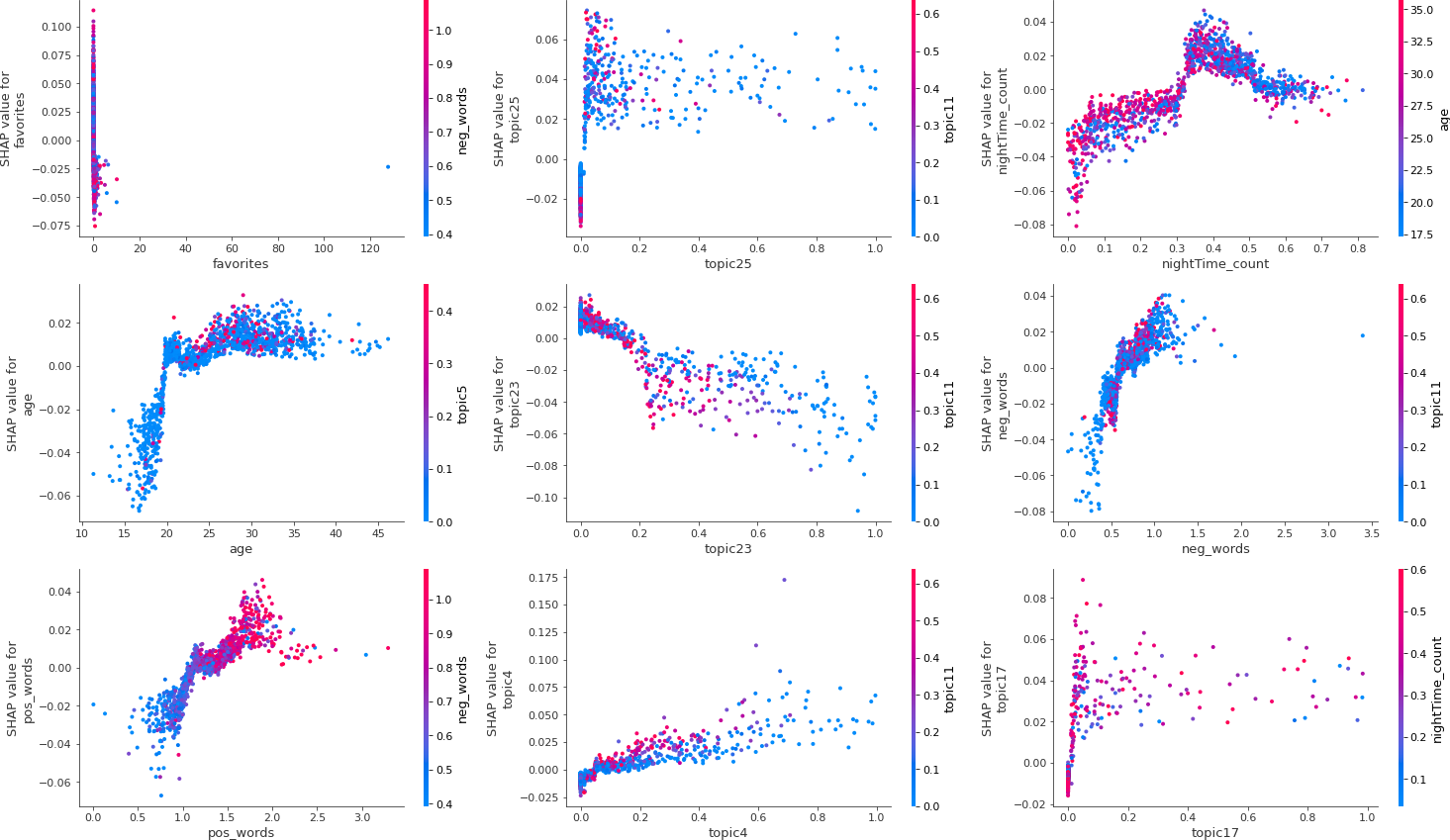}
         \caption{CLPsych 2015 CvP}
		
     \end{subfigure}
     \hfill
     \begin{subfigure}[b]{1\textwidth}
         \centering
         \includegraphics[width=1\textwidth]{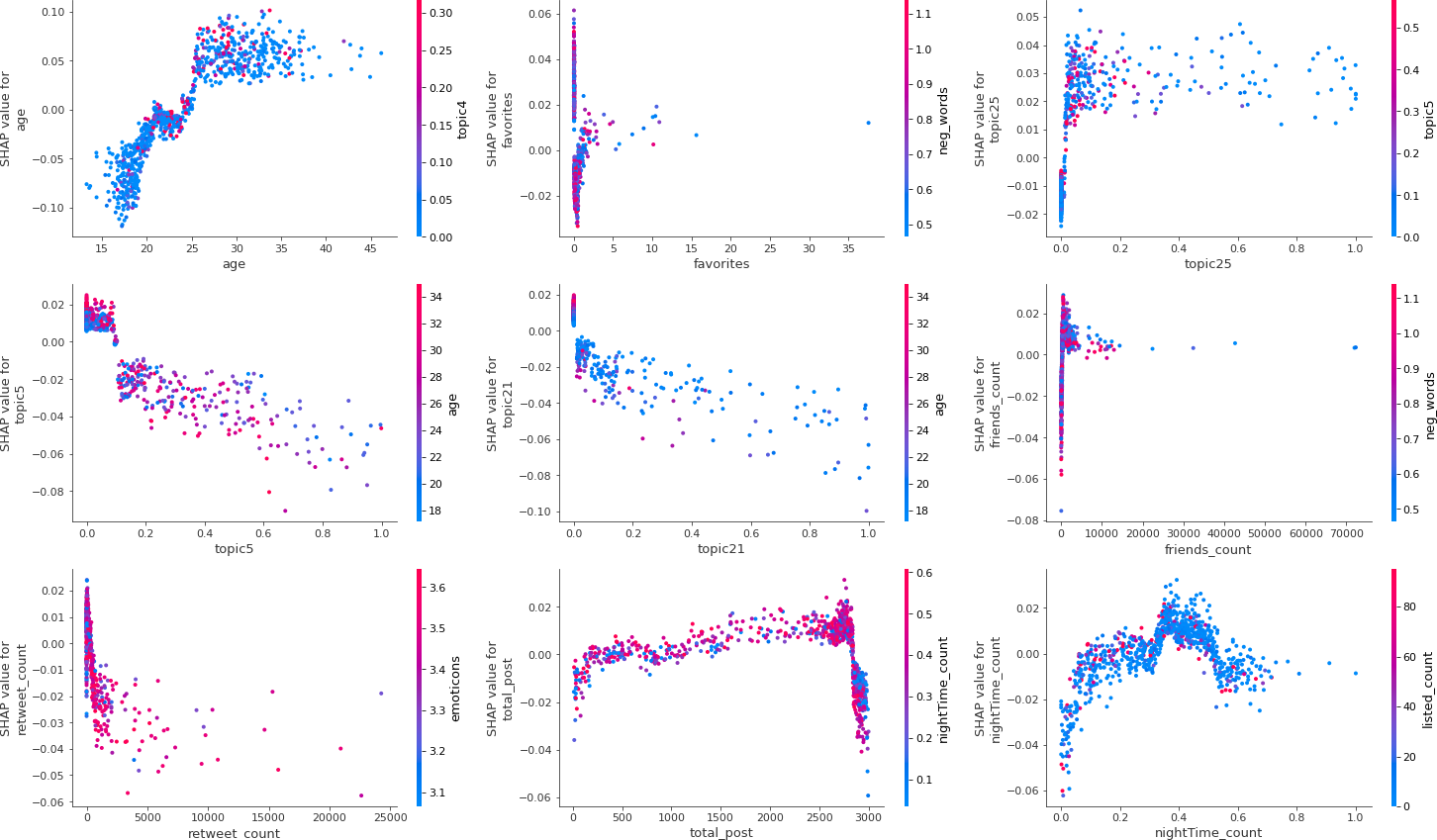}
         \caption{CLPsych 2015 PvD subset}
     \end{subfigure}
     \hfill

        \caption{Top 9 significant feature dependency for CLPsych 2015 CvP and PvD subsets.}
        \label{fig:dp_supply}
\end{figure*}

Fig. \ref{fig:feature_importances2}, \ref{fig:dp_supply} shows the feature importance and dependency plots for the subsets of CLPsych 2015, CvP and PvD.
\begin{table*}[htbp]
\caption{LDA Topics of the TTDD dataset}
\label{Tab:lda_ttdd}
\centering
\begin{tabular}{|l|l|l|}
\hline
\textbf{Feature Name} &\multicolumn{1}{l|}{\textbf{Deduced Topic}}  & \multicolumn{1}{c|}{\textbf{Top 10 words}} \\ \hline
Topic1      &Youtube video       & 'earning', 'mplusrewards', 'stream', 'review', 'link', 'youtube', 'android', 'download', 'iphone', 'awesome'                \\ \hline
Topic2      &Politics       & 'trump', 'obama', 'russia', 'president', 'russian', 'america', 'election', 'putin', 'donald', 'vote'                        \\ \hline
Topic3      &Harry Potter       & 'woke', 'onair', 'john', 'david', 'drew', 'service', 'brah', 'potter', 'harry', 'magic'                                     \\ \hline
Topic4      &India       & 'india', 'dance', 'beach', 'modi', 'download', 'austin', 'ballad', 'loudingh', 'indian', 'hit'                              \\ \hline
Topic5       &Band      & 'radio', 'funk', 'disco', 'bigbang', 'teamgot7', 'lovelyz', 'got7', 'ikon', 'dara', 'album'                                 \\ \hline
Topic6        &Policy     & 'wood', 'december', 'january', 'country', 'business', 'police', 'power', 'issue', 'service', 'case'                         \\ \hline
Topic7      &Facebook       & 'wearing', 'posted', 'facebook', 'gratitude', 'happiness', 'shoutout', 'computer', 'friedrich', 'mumbai', 'dear'            \\ \hline
Topic8       &Film      & 'carrie', 'wholesome', 'fisher', 'record', 'soul', 'war', 'film', 'reynolds', 'goldenglobes', 'character'                   \\ \hline
Topic9       &Contest      & 'giveaway', 'enter', 'entered', 'vote', 'competition', 'contest', 'model', 'voted', 'blog', 'teen'                          \\ \hline
Topic10       &Album     & 'jimin', 'yoongi', 'stan', 'jungkook', 'album', 'group', 'taehyung', 'dance', 'member', 'stage'                             \\ \hline
Topic11      &Online Music      & 'nowplaying', 'listenlive', 'online', 'grayson', 'dolantwinsnewvideo', 'dolan', 'island', 'paradise', 'gameinsight', 'link' \\ \hline
Topic12       &Abusive language     & 'nigga', 'tryna', 'bout', 'yall', 'lmfao', 'dick', 'hoe', 'bruh', 'female', 'ugly'                                          \\ \hline
Topic13    &Greeting        & 'haha', 'hahaha', 'lovely', 'merry', 'goodnight', 'awesome', 'babe', 'band', 'sherlock', 'listening'                        \\ \hline
Topic14     &Website       & 'click', 'website', 'profile', 'invite', 'adult', 'positive', 'york', 'link', 'cost', 'porn                                 \\ \hline
Topic15      &Mental health      & 'mental', 'health', 'anxiety', 'mentalhealth', 'fear', 'bellletstalk', 'loving', 'illness', 'happiness', 'article'          \\ \hline
Topic16     &Illness       & 'bipolar', 'cancer', 'online', 'disorder', 'doctor', 'treatment', 'injury', 'syria', 'symptom', 'latest'                    \\ \hline
Topic17      &Radio      & 'playlist', 'retweeted', 'tune', 'feat', 'ruby', 'radio', 'dirty', '1xtra', 'official', 'hiphop'                            \\ \hline
Topic18     & Animation      & 'character', 'anime', 'draw', 'holy', 'meme', 'drawing', 'forgot', 'entire', 'artist', 'episode'                            \\ \hline
Topic19      & Music     & 'zayn', 'iheartawards', 'bestmusicvideo', 'pillowtalk', 'vote', 'voting', 'radio', 'now2016', 'taylor', 'nowzayn'           \\ \hline
Topic20      & $\sim$      & 'jasmin', 'fuckin', 'luke', 'event', 'sheskindahotvma', 'screen', '5sos', 'card', 'noctis', 'ffxv'                          \\ \hline
Topic21       &Interest     & 'justin', 'pokemon', 'hack', 'teamfollowback', 'niall', 'harry', 'followback', 'bieber', 'louis', 'direction'               \\ \hline
Topic22       &Fan     & '9gag', 'album', 'nigga', 'player', 'trash', 'brown', 'bruh', 'fan', 'ugly', 'soulja'                                       \\ \hline
Topic23       &Premier League     & 'player', 'league', 'nigerian', 'arsenal', 'chelsea', 'united', 'fan', 'club', 'nigeria', 'mate'                            \\ \hline
Topic24       &Follower     & 'episode', 'walking', 'rick', 'daryl', 'follower', 'checked', 'stats', 'automatically', 'unfollowers', 'unfollowed'         \\ \hline
Topic25       &Describing People     & 'soul', 'funk', 'motown', 'kiss', 'university', 'personal', 'softly', 'slowly', 'hug', 'lip'                                \\ \hline
\end{tabular}
\end{table*}

\begin{table*}[htbp]
\caption{LDA Topics of the CLPsych 2015 dataset}
\label{Tab:lda_clp}
\centering
\begin{tabular}{|l|l|l|}
\hline
\textbf{Feature Name} & \multicolumn{1}{l|}{\textbf{Deduced Topic}}                                                                                                   & \multicolumn{1}{c|}{\textbf{Top 10 words}}                                                                                                   \\ \hline
Topic1    &Name         & 'luke', '5sos', 'hemmings', 'glee', 'peopleschoice', 'blaine', '5sosfam', 'penguin', 'kurt', 'ashton'                               \\ \hline
Topic2      &Lottery       & 'reward', 'earning', 'mpoints', 'hiking', 'camping', 'hcsm', 'riley', 'trail', 'scot', 'getglue'                                    \\ \hline
Topic3       &Travel      & 'photoset', 'neuro', 'toronto', 'giveaway', 'musical', 'theatre', 'broadway', 'fallow', 'ford', 'canadian'                          \\ \hline
Topic4         &Mental health    & 'mentalhealth', 'vitamin', 'ptsd', 'medical', 'ferguson', 'ebola', 'auspol', 'republican', 'abbott', 'veteran'                      \\ \hline
Topic5       &Mobile games      & 'gameinsight', 'ipad', 'ipadgames', 'coin', 'collected', 'android', 'androidgames', 'harvested', 'tribez', 'turk'                   \\ \hline
Topic6        &People     & 'niall', 'liam', 'zayn', 'louis', 'mtvhottest', 'fandom', 'aries', 'luke', 'ilysm', 'icon'                                          \\ \hline
Topic7      &Salutation       & 'capricorn', 'tryna', 'bruh', 'oomf', 'hella', 'dope', 'homie', 'blunt', 'lmaooo', 'lmaoo'                                          \\ \hline
Topic8      &Religion       & 'grace', 'bible', 'praise', 'bishop', 'psalm', 'glory', 'worship', 'cont', 'pastor', 'sfgiants'                                     \\ \hline
Topic9      &Racing      & 'demi', '2day', 'nascar', 'calorie', 'brianna', 'twitition', 'racing', 'cont', 'disorder', 'thinspo'                                \\ \hline
Topic10      &$\sim$    & 'cunt', 'hockey', 'realise', 'potter', 'drum', 'dunno', 'mum', 'colour', 'genuinely', 'australia'                                   \\ \hline
Topic11    &News        & 'pisces', 'austin', 'voteaustinmahone', 'entrepreneur', 'emazing', 'startup', 'goodmorning', 'beliebers', 'dallas', 'palace'        \\ \hline
Topic12      &$\sim$      & 'subscribe', 'tcot', 'pjnet', 'katy', 'columbus', 'isi', 'perry', 'veteran', 'obamacare', 'american'                                \\ \hline
Topic13      &Scotland      & 'dundee', 'salary', 'fife', 'scotland', 'cornwall', 'getglue', 'engineer', 'royal', 'negotiable', 'kris'                            \\ \hline
Topic14       &Middle East Countries     & 'syria', 'contacted', 'producer', 'package', 'balochistan', 'domestic', 'gaming', 'terrorism', 'soundcloud', 'pokemon'              \\ \hline
Topic15       &Feminism      & 'virgo', 'reckless', 'mixtape', 'fringe', 'tiffany', 'feminist', 'violence', 'hbic', 'gender', 'feminism'                           \\ \hline
Topic16      &Museum      & 'illustration', 'recipe', 'webdesign', 'vintage', 'china', 'museum', 'data', 'hubby', 'writer', 'silver'                            \\ \hline
Topic17       &Government   & 'china', 'tory', 'paedophile', 'minister', 'stylish', 'govt', 'labour', 'cameron', 'scotland', 'injured'                            \\ \hline
Topic18     &$\sim$       & \begin{tabular}[c]{@{}l@{}}'auction', 'vietnam', 'xxxx', 'freemarinea', 'marine', 'ahaha', 'thankyou', 'kristen', 'xxxxx',\\ 'awkwardshoppingsituations' \end{tabular}                                                                                                                                        \\ \hline
Topic19    &Friend       & 'hahah', 'sagittarius', 'bestfriend', 'soccer', 'gunna', 'oomf', 'freshman', 'semester', 'lolol', 'prom'                            \\ \hline
Topic20     &$\sim$       & 'jasmin', 'fuckin', 'luke', 'event', 'sheskindahotvma', 'screen', '5sos', 'card', 'noctis', 'ffxv'                                  \\ \hline
Topic21      &Receipt      & 'inbox', 'mobile', 'marketing', 'consumer', 'hosting', 'connect', 'maria', 'device', 'monmouth', 'sandy'                            \\ \hline
Topic22      &Autism      & 'amas', 'phillip', 'bpdchat', 'phillips', 'ptsd', 'autism', 'autistic', 'mhchat', 'maternal', 'trauma'                              \\ \hline
Topic23      &Region      & 'wifey', 'israel', 'baseball', 'league', 'gaza', 'defense', 'boston', 'anonymous', 'playoff', 'tiger'                               \\ \hline
Topic24      &Social Media      &\begin{tabular}[c]{@{}l@{}} 'stats', 'unfollowers', 'unfollower', 'highered', 'teamfollowback', 'submission', 'mixtape', 'unfollowed', 'grilling',\\ 'followback'\end{tabular}                                                                                                                                  \\ \hline
Topic25      &$\sim$        &  'peopleschoice', 'votearianagrande', 'castle', 'taurus', 'libra', 'demi', 'oomf', 'tryna', 'lovato', 'goodmorning'                  \\ \hline

\end{tabular}
\end{table*}

\begin{figure*}[htbp]
     \centering
     \begin{subfigure}[b]{0.48\textwidth}
         \centering
         \includegraphics[width=1\textwidth]{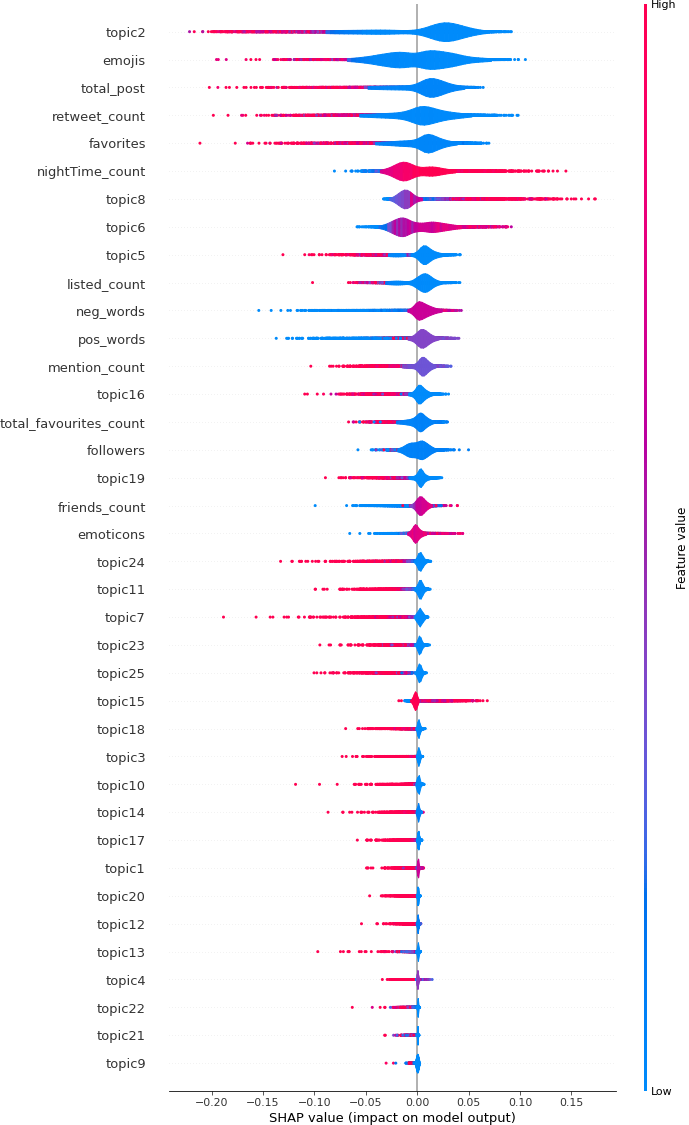}
         \caption{TTDD: Impact of each feature on the model prediction.}
		
     \end{subfigure}
     \hfill
     \begin{subfigure}[b]{0.48\textwidth}
         \centering
         \includegraphics[width=1\textwidth]{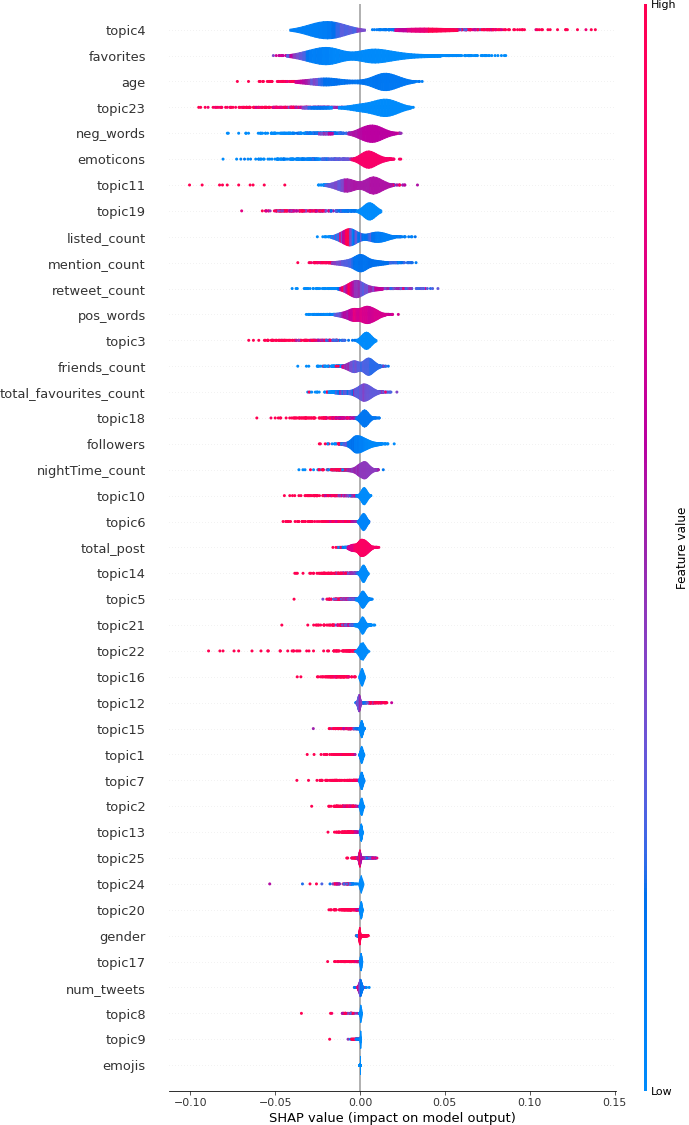}
         \caption{CLPsych 2015: Impact of each feature on the model prediction.}
     \end{subfigure}
     \hfill

        \caption{Feature Importance.}
        \label{fig:feature_importances}
\end{figure*}

\newpage

\newpage
\section{Supplementary D: TreeSHAP}
\subsection{Introduction of the TreeSHAP}
TreeSHAP \cite{lundberg2020local} is an algorithm to compute Shapley Values \cite{shapley1951notes} for Decision Trees based models. The goal of TreeSHAP is to explain the prediction for an instance with its features' contributions. The Shapley Value of a feature is its contribution to the prediction. The Shapley Value is weighted and summed over all possible feature combinations:
\begin{equation}
\label{eqn:shapley}
\phi_{X_{i}^{v}}(f) = \sum_{S\subseteq \left \{ X^{1}_{i}, ..., X^{V}_{i}  \right \}\setminus \left \{ X^{v}_{i} \right \}} \frac{|S|!(V-|S|-1)!}{V!}\left [f(S\cup \left \{ X^{v}_{i} \right \}) - f(S) \right ]
\end{equation}
where $V$ is the number of features and $X_{i}$ is a data instance. $S$ represents the feature subset which does not include feature $X^{v}_{i} \in \mathbb{R}^{N \times V}$ ($N$ is the number of the data instances) and $f()$ is the prediction function for the machine learning model. TreeSHAP describes the following three desirable properties:
\begin{enumerate}[(1)]

\item {Local Accuracy: 
\begin{equation}
\label{eqn:local_accuracy}
f(X_{i})=\phi_{0}(f) +\sum_{v=1}^{V}\phi_{X_{i}^{v}}(f) 
\end{equation}
where $\phi_{0}(f)$ is the model's expectation:
\begin{equation}
\label{eqn:expected_value}
\phi_{0}(f) = \frac{N_{target}}{N}
\end{equation}
$N$ is the number of the data instances and $N_{target}$ is the number of the target class samples (e.g. depressed users). In Eq. \ref{eqn:local_accuracy}, $f(X_{i})$ is the model output for instance $X_{i}$ and the sum of the features' Shapley Values $\phi_{X_{i}^{v}}(f) $ matches the change of the model output $f(X_{i})-\phi_{0}(f)$.
}

\item {Consistency: For any two models $f()$ and $f^{'}()$, if 
\begin{equation}
\label{eqn:consistency}
f^{'}(S\cup \left \{ X_{i}^{v} \right \}) - f^{'}(S) \geq f(S\cup \left \{ X_{i}^{v} \right \}) - f(S)
\end{equation}
For all feature subsets $S\in X_{i}$, $\phi_{X_{i}^{v}}(f^{'}) \geq \phi_{X_{i}^{v}}(f)$. If a model changes so that the marginal contribution of a feature increases or maintains, the Shapley Value also increases or maintains.
}
\item {Missingness: If 
\begin{equation}
\label{eqn:missingness}
f(S\cup \left \{ X_{i}^{v} \right \}) = f(S)
\end{equation}
For all feature subsets $S\in X_{i}$, $\phi_{X_{i}^{v}}(f)=0$. Feature $X_{i}^{v}$ does not have any impact on the model $f()$.
}
\end{enumerate}
To conclude, $\phi_{X_{i}^{v}}(f)$ is the Shapley Value or the feature contribution to the model prediction. Using our depression detection dataset as an example, the features with positive Shapley Values increase the predicted depression risks and vice versa. 

Besides, TreeSHAP also allows us to compute interaction depth by considering pairwise feature attributions. The interaction depth is based on the Pearson Product-Moment Correlation \cite{rupinski1996approximating} and is given by:
\begin{equation}
\label{eqn:interaction_effect}
\rho_{v,j}  =  \frac{\sum_{i=1}^{N}\left [ \phi_{X_{i}^{v}}(f)-\overline{\phi_{X^{v}}} \right]\left [ X_{i}^{j}-\overline{X^{j}} \right ]}{\sqrt{\sum_{i=1}^{N}\phi_{X_{i}^{v}}(f)-\overline{\phi_{X^{v}}}}\sqrt{\sum_{i=1}^{N}X_{i}^{j}-\overline{X^{j}}} }
\end{equation}
where $X^{v}$ is the values of feature $X^{v}_{i}$ for all the examples in a dataset. $\rho_{v,j}$ is the correlation coefficient of features $X^{v}$ with features $X^{j}$ (s.t. $v \neq  j \& \rho_{v,j} \neq \rho_{j,v}$). $\overline{\phi_{X^{v}}}$ is the mean of Shapley Values against features $X^{v}$ and $\overline{X^{j}}$ is the mean of features $X^{j}$. For features $X^{v}$, Eq. (6) will allow us to calculate the correlation value of $\phi_{X^{v}}(f)$ with the other features' values.  The feature with the highest correlation value will be regarded as the most interactive feature attribution of $X^{v}$.

\subsection{Example of the Shapley Value Calculation}
In order to explain how TreeSHAP method is integrated with our proposed CBPT model, we give an example to simply calculate the features’ Shapley Values. 

\begin{figure*}
     \centering
         \includegraphics[width=1\textwidth]{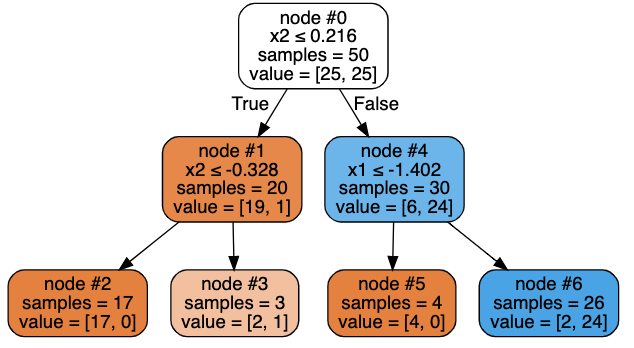}
         \caption{A decision tree example. In each tree node (rectangle),  "$node \#0$" is the id of the tree node. "$X^{v}$" is the used feature for splitting instances. If the corresponding feature values of a instance is smaller than $X^{v}$, the instance will be send to the left children node otherwise to the right children node. "samples=50" represents the total number of the instances in tree node "$node \#0$". "value=[25,25]" represents the number of different classes' instances. The first 25 in the value list represents 25 samples of class 0 and the second one is the number of class 1 samples.}
         
        \label{fig:treeforshap}
\end{figure*}

Fig. \ref{fig:treeforshap} is a two-depth decision tree which is trained using 50 random instances generated by the Scikit-learn package \cite{pedregosa2011scikit}. Here, the generated training set is balanced and includes two classes' data instances with three feature dimensions. Let's compute the Shapley Values for an instance $X_{i}=[1,0,1]$. We define $X_{i}^{v}$ as the value of the $v$-th feature for instance $X_{i}$. We use $N_{d}$ to denote the number of the instances with $node\#d$.

Instance $X_{i}$ has 3 independent features, so we have to consider $3!=6$ feature combinations. Marginal contribution will represent a local Shapley Value of a feature in each combination. Note that TreeSHAP makes the assumption (Eq. \eqref{eqn:local_accuracy}) that the model prediction for instance $X_{i}$ is equal to the sum of its features' Shapley Values plus the model's expectation. Assuming class 1 is the target class, the expectation of the decision tree is available using Eq. \eqref{eqn:expected_value}:

\begin{equation}
\label{eqn:example1}
\phi_{0}(f)  =f(S) = \frac{N_{target}}{N} = \frac{25}{50} = \frac{1}{2} 
\end{equation}
\textbf{Then, consider the feature combination $X^{1}_{i}>X^{2}_{i}>X^{3}_{i}$:} 
\begin{enumerate}[(1)]

\item {First, let feature $X^{1}_{i}$ be available, and the values of $X^{2}_{i}$ and $X^{3}_{i}$ are null temporarily. The first node $node\#0$ uses $X^{2}$ as the split variable. Since $X^{2}_{i}$ is not available yet, the prediction is computed according to the decision tree theory $\left[\frac{20}{50}\times (\text{prediction for the left child node}\#1)+ \frac{30}{50}\times (\text{prediction for the right child node}\#4)\right]$.
\begin{enumerate}[(a)]
\item {Prediction for the left child $node\#1$: $node\#1$ still uses $X^{2}$ as the split variable. Similarly, $X^{2}_{i}$ is not available yet, the prediction for $node\#1$ is computed:
\begin{equation}
\label{eqn:example2}
\frac{17}{20}\times (\text{prediction for the left child node}\#2)+ \frac{3}{20}\times (\text{prediction for the right child node}\#3)
\end{equation}
For $node\#2$, the model prediction for class 1 is $\frac{0}{17+0}=0$ and $\frac{1}{1+2}=\frac{1}{3}$ is the result for $node\#3$. So the prediction result for $node\#1$ is $\frac{17}{20} \times 0 +\frac{3}{20} \times \frac{1}{3} = \frac{1}{20} $. 

}
\item {Prediction for the right child $node\#4$: $node\#4$ uses $X^{1}$ as the split variable, the prediction for $node\#4 =\text{the result for }node\#6 =\frac{24}{26}$.}
\end{enumerate}
Hence, the prediction for the instance with only feature $X^{1}_{i}$ is $\left[\frac{20}{50} \times \frac{1}{20}  + \frac{30}{50} \times \frac{24}{26}  \right]  \approx 0.574$. The marginal contribution of $X^{1}_{i}$ in this combination is:
\begin{equation}
\label{eqn:example3}
\phi_{X^{1}_{i}}^{1} =f(S\cup \left \{ X^{1}_{i} \right \}) - f(S)  =0.574 - \phi_{0}(f) = 0.074
\end{equation}
where superscript $1$ of $\phi_{X^{1}_{i}}^{1}$ is the index of  the feature combination.
}
\item{Then, we let feature $X^{2}_{i}$ be available.  Instance $X_{i}$ can pass through $node\#0$ and $node\#1$ finally land at $node\#3$. The marginal contribution of $X^{2}_{i}$ in this combination is:
\begin{equation}
\label{eqn:example4}
\phi_{X^{2}_{i}}^{1} =  f(S\cup \left\{X^{1}_{i},X^{2}_{i}\right\} ) -f(S\cup \left\{X^{1}_{i}\right\} ) = \frac{1}{3} - 0.574 \approx -0.241
\end{equation}
}
\item{Finally, we let feature $X^{3}_{i}$ be available. Because $X^{3}_{i}$ is not used as a split variable in any of the tree nodes, adding this feature does not alter the model prediction in any way. According to Eq. \eqref{eqn:missingness}, we have $\phi_{X^{3}_{i}}^{1} =0$. }
\end{enumerate}
\textbf{Next, consider the combination $X^{2}_{i}>X^{1}_{i}>X^{3}_{i}$:} 
\begin{enumerate}[(1)]
\item{In this feature combination, feature $X^{2}_{i}$ is available firstly. $node\#0$ and $node\#1$ use $X^{2}$ as their split variable. Instance $X_{i}$ can pass through $node\#0$ and $node\#1$ finally reach $node\#3$. The marginal contribution of $X^{2}_{i}$ is:
\begin{equation}
\label{eqn:example5}
\phi_{X^{2}_{i}}^{2}= \text{Prediction for }node\#3 - \phi_{0}(f) = \frac{1}{3} - \frac{1}{2} \approx -0.167
\end{equation}
}
\item{Then, we let feature $X^{1}_{i}$ be available. Since adding $X^{1}_{i}$ does not alter the prediction result for instance $X_{i}$, the marginal contribution $\phi_{X^{1}_{i}}^{2} = 0$ (Eq. \eqref{eqn:missingness}). }
\item{Finally, we let feature $X^{3}_{i}$ be available. Therefore, the marginal contribution of feature $X^{3}_{i}$ is $\phi_{X^{3}_{i}}^{2} = 0$ (Eq. \eqref{eqn:missingness}). }
\end{enumerate}
Similarly, the marginal contributions of the remaining feature combinations are shown as follows:
\\
\textbf{Combination $X^{1}_{i}>X^{3}_{i}>X^{2}_{i}$:} $\phi_{X^{1}_{i}}^{3}=0.074,\phi_{X^{2}_{i}}^{3}=-0.241,\phi_{X^{3}_{i}}^{3}=0$
\\
\textbf{Combination $X^{2}_{i}>X^{3}_{i}>X^{1}_{i}$:} $\phi_{X^{1}_{i}}^{4}=0,\phi_{X^{2}_{i}}^{4}=-0.167,\phi_{X^{3}_{i}}^{4}=0$
\\
\textbf{Combination $X^{3}_{i}>X^{1}_{i}>X^{2}_{i}$:} $\phi_{X^{1}_{i}}^{5}=0.074,\phi_{X^{2}_{i}}^{5}=-0.241,\phi_{X^{3}_{i}}^{5}=0$
\\
\textbf{Combination $X^{3}_{i}>X^{2}_{i}>X^{1}_{i}$:} $\phi_{X^{1}_{i}}^{6}=0,\phi_{X^{2}_{i}}^{6}=-0.167,\phi_{X^{3}_{i}}^{6}=0$

Therefore, the Shapley Values for instance $X_{i}$ are computed:
\begin{equation}
\label{eqn:example6}
\begin{split}
&\phi_{X^{1}_{i}}(f)=\frac{\sum_{c=1}^{3!} \phi_{X^{1}_{i}}^{c}} { 3!}  = 0.037 \\
&\phi_{X^2_{i}}(f)=\frac{\sum_{c=1}^{3!} \phi_{X^{2}_{i}}^{c}} { 3!}  = -0.204 \\
&\phi_{X^{3}_{i}}(f)=\frac{\sum_{c=1}^{3!} \phi_{X^{3}_{i}}^{c}} { 3!}  = 0 \\
\end{split}
\end{equation}

From the decision tree, we observe that the predicted probability of instance $X_{i}$ for class 1 is $f(X_{i})=\frac{1}{3}$. Using Eq. \eqref{eqn:local_accuracy}, we have:
\begin{equation}
\label{eqn:example7}
 f(X_{i})=\phi_{0}(f) +\sum_{v=1}^{V}\phi_{X_{i}^{v}}(f)  =  \frac{1}{2} + (0.037-0.204) = 0.333 \approx \frac{1}{3}
\end{equation}
That means feature $X^{1}_{i}$ increases the predicted probability of class 1 for instance $X_{i}$ by 3.7\%. Feature $X_{i}^{2}$ decreases the predicted probability of class 1 by 20.4\%. From these values, we understand which feature affects the model's prediction result. 

Finally, note that our proposed CBPT is an ensemble method which includes multiple decision tree based estimators. A single feature's contribution to the prediction of CBPT is computed:
\begin{equation}
\label{eqn:example8}
\phi_{X_{i}^{v}}(f)=\frac{\sum_{k=1}^{K}\theta_{k}\phi_{X_{i}^{v} }(f_{k})} {K}
\end{equation}
where $K$ is the number of the pruned trees and $\theta_{k}$ is the weight of the corresponding estimator.

\end{document}